\documentclass{article}

\usepackage{times}
\usepackage{graphicx} 

\usepackage{natbib}

\usepackage{algorithm}
\usepackage{algorithmic}
\usepackage{graphicx}
\usepackage{amsfonts}
\usepackage{amsmath}
\usepackage{amsthm}
\usepackage{booktabs}
\usepackage[percent]{overpic}
\usepackage{csquotes}
\usepackage{color}
\usepackage{caption}
\usepackage{subcaption}
\usepackage{xspace}
\usepackage{verbatim}
\usepackage{multirow}

\newtheorem{theorem}{Theorem}[section]
\newtheorem{lemma}[theorem]{Lemma}

\usepackage[accepted]{icml2016}

\icmltitlerunning{Differential Imitation Learning for Sequential Prediction}

\begin{document} 

\twocolumn[



\icmltitle{Deeply AggreVaTeD:\\ Differentiable Imitation Learning for Sequential Prediction}


\icmlauthor{Wen Sun$^{\dagger}$}{wensun@cs.cmu.edu}
\icmlauthor{Arun Venkatraman$^{\dagger}$}{arunvenk@cs.cmu.edu}
\icmlauthor{Geoffrey J. Gordon$^{\dagger}$}{ggordon@cs.cmu.edu}
\icmlauthor{Byron Boots$^*$}{bboots@cc.gatech.edu}
\icmlauthor{J. Andrew Bagnell$^{\dagger}$}{dbagnell@ri.cmu.edu}
\icmladdress{$^\dagger$School of Computer Science, Carnegie Mellon University, USA \\
$^*$College of Computing, Georgia Institute of Technology, USA 
}
\icmlkeywords{imitation learning, sequential prediction, reinforcement learning, sequential prediction}

\vskip 0.3in
]



\begin{abstract}
Researchers have demonstrated state-of-the-art performance in sequential decision making problems (e.g., robotics control, sequential prediction) with deep neural network models.
One often has access to near-optimal oracles that achieve good performance on the task during training.  We demonstrate that \textit{AggreVaTeD} --- a policy gradient extension of the Imitation Learning (IL) approach of \cite{ross2014reinforcement} --- can leverage such an oracle to achieve faster and better solutions with less training data than a less-informed Reinforcement Learning (RL) technique. Using both feedforward and recurrent neural predictors, we present stochastic gradient procedures on a sequential prediction task, dependency-parsing from raw image data, as well as on various high dimensional robotics control problems. We also provide a comprehensive theoretical study of IL that demonstrates we can expect up to \textit{exponentially} lower sample complexity for learning with \textit{AggreVaTeD} than with RL algorithms, which backs our empirical findings. Our results and theory indicate that the proposed approach can achieve superior performance with respect to the oracle when the demonstrator is sub-optimal.


\end{abstract}



\section{Introduction}
A fundamental challenge in artificial intelligence, robotics, and language processing is to reason, plan, and make a sequence of decisions to minimize accumulated cost, achieve a long-term goal, or optimize for a loss acquired only after many predictions. Reinforcement Learning (RL), especially deep RL, has dramatically advanced the state of the art in sequential decision making in high-dimensional robotics control tasks as well as in playing video and board games~\cite{schulman2015trust,silver2016mastering}. Though conventional supervised learning of deep models has been pivotal in advancing performance in sequential prediction problems, researchers are beginning to utilize deep RL methods to achieve high performance~\cite{ranzato2015sequence,bahdanau2016actor,li2016deep}. Often in sequential prediction tasks, future predictions from the learner are dependent on the history of previous predictions; thus, a poor prediction early on can yield high accumulated loss (cost) for future predictions. Viewing the predictor as a \emph{policy} $\pi$, deep RL algorithms are able to reason about the future accumulated cost in a sequential decision making process whether it is a traditional robotics control problem or a sequential structured prediction task.

In contrast with reinforcement learning methods, well-known \emph{imitation learning} (IL) and sequential prediction algorithms such as SEARN \cite{daume2009search}, DaD \cite{venkatraman2015improving}, AggreVaTe \cite{ross2014reinforcement}, and LOLS \cite{chang2015learning} reduce the sequence prediction problem to supervised learning by leveraging one special property of the sequence prediction problem: at training time we usually have a (near) optimal \emph{cost-to-go oracle}. At any point along the sequential prediction process (i.e. state or partially completed sequential prediction), the oracle is able to select the next (near)-best action. 

Concretely, the above methods assume access to an oracle\footnote{Expert, demonstrator, and oracle are used interchangeably.} that provides an optimal or near-optimal action and the future accumulated loss $Q^*$, also called the cost-to-go. For robotics control problems, this oracle may come from a human expert guiding the robot during the training phase~ \cite{abbeel2004apprenticeship} or from an optimal MDP solver~\cite{Ross2011_AISTATS,kahn2016plato} that either may be too slow to use at test time or leverages information unavailable at test time (e.g., ground truth). Similarly, for sequential prediction problems, an oracle can be constructed by optimization (e.g., beam search) or by a clairvoyant greedy algorithm~\cite{daume2009search,ross2013contextual,rhinehart2015visual,chang2015learning_dependency} that is near-optimal on the task specific performance metric (e.g., cumulative reward, IoU, Unlabeled Attachment Score, BLEU) given the training data's ground truth.

Such an oracle, however, is only available during training time (e.g., when there is access to ground truth). Thus, the goal of IL is to learn a policy $\hat \pi$, with the help of the oracle $(\pi^*, Q^*)$ during the training session, such that $\hat \pi$ achieves similar quality performance at test time when the oracle is unavailable. In contrast to IL, reinforcement learning methods often initialize with an random policy $\pi_0$ or cost-to-go (accumulated loss) $Q_0$ predictor which may be far from the optimal. The optimal policy (or cost-to-go) must be found through a tradeoff of dithering, directed exploration, and exploitation.

The existence of oracle can be exploited to alleviate blind learning by trial and error: one can \emph{imitate} the oracle to speed up learning process by significantly reducing exploration. A classic IL method is to collect data from running the demonstrator or oracle and train a regressor or classifier via supervised learning. These methods \cite{abbeel2004apprenticeship,syed2008apprenticeship,ratliff2006maximum,ziebart2008maximum,finn2016guided,ho2016generative} learn either a policy $\hat \pi^*$ or $\hat Q^*$ from a fixed-size dataset pre-collected from the oracle. A pernicious problem with these  methods is that they require the training and test data to be sampled from the same distribution.  This is very difficult to enforce in practice, and, as a result, policies learned by these methods can fail spectacularly in theory and in practice~\cite{ross2010efficient}. \emph{Interactive} approaches to IL such as SEARN \cite{daume2009search}, DAgger \cite{Ross2011_AISTATS}, and AggreVaTe \cite{ross2014reinforcement} interleave learning and testing procedures to overcome the data mismatch issue and, as a result, work well in practical applications. Furthermore, these interactive approaches can provide strong theoretical guarantees between training time loss and test time performance through a reduction to no-regret online learning.

In this work, we introduce \emph{AggreVaTeD}, a \emph{differentiable} version of AggreVaTe (Aggregate Values to Imitate \cite{ross2014reinforcement}) which extends interactive IL for use in sequential prediction and challenging continuous robot control tasks.  We provide two gradient update procedures: a regular gradient update developed from Online Gradient Descent (OGD) \cite{Zinkevich2003_ICML} and a natural gradient update  \cite{kakade2002natural,bagnell2003covariant} which we show is closely related to Exponential Gradient Descent (EG), another popular no-regret algorithm that enjoys an almost dimension-free property \cite{bubeck2015convex}.\footnote{i.e., the regret bound depends on poly-log of the dimension of parameter space.} 

AggreVaTeD leverages the oracle to learn rich polices that can be represented by complicated non-linear function approximators. Our experiments with deep neural networks on various robotics control simulators and on a dependency parsing sequential prediction task show that AggreVaTeD can achieve expert-level performance and even \emph{super-expert} performance when the oracle is sub-optimal, a result rarely achieved by non-interactive IL approaches. The differentiable nature of AggreVaTeD additionally allows us to employ LSTM-based policies to handle partially observable settings (e.g., observe only partial robot state). Empirical results demonstrate that by leveraging an oracle, IL can learn much faster than RL in practice.


In addition to providing a set of practical algorithms, we develop a comprehensive theoretical study of IL on discrete MDPs. We construct an MDP that demonstrates exponentially better sample efficiency for IL than any RL algorithm. For general discrete MDPs, we provide a regret upper bound for AggreVaTeD with EG, which shows IL can learn dramatically faster than RL. We provide a regret lower bound for any IL algorithm, which demonstrates that AggreVaTeD with EG is near-optimal. Our experimental and the theoretical results support the proposition: 
\begin{displayquote}
Imitation Learning is a more effective strategy than Reinforcement Learning for sequential prediction with near optimal cost-to-go oracles.
\end{displayquote}

\section{Preliminaries}
In order to develop algorithms that reason about long-term decision making, it is convenient to cast the problem into the Markov Decision Process (MDP) framework. The MDP framework consists of a set of states, actions (that come from a policy), cost (loss), and a model that transitions states given actions. For robotics control problems, the robot's configuration is the state, the controls (e.g., joint torques) are the actions, and the cost is related to achieving a task (e.g., distance walked). Though more nuanced, most sequential predictions can be cast into the same framework~\cite{daume2009search}. The actions are the learner's (e.g., RNN's) predictions. The state is then the result of all the predictions made so far (e.g., the dependency tree constructed so far or the words translated so far). The cumulative cost is the performance metric such as (negative) UAS, received at the end (horizon) after the final prediction.

Formally, a finite-horizon Markov Decision Process (MDP) is defined as $(\mathcal{S}, \mathcal{A}, P, C, \rho_0, H)$. Here, $\mathcal{S}$ is a set of $S$ many states and $\mathcal{A}$ is a set of $A$ actions; given time step $t$, $P_t$ is the transition dynamics such that for any $s_t\in\mathcal{S},s_{t+1}\in\mathcal{S}, a_t\in\mathcal{A}$, $P_t(s_{t+1}|s_t, a_t)$ is the probability of transiting to state $s_{t+1}$ from state $s_t$ by taking action $a_t$ at step $t$; $C$ is the cost distribution such that a cost $c_t$ at step $t$ is sampled from $C_t(\cdot | s_t, a_t)$. Finally, we denote $\bar{c}_t$ as the expected cost, $\rho_0$ as the initial distribution of states, and $H\in\mathbb{N}^+$ as the finite horizon (max length) of the MDP. 

We define a stochastic policy $\pi$ such that for any state $s\in\mathcal{S}$, $\pi(\cdot | s)\in \Delta(A)$, where $\Delta(A)$ is a $A$-dimension simplex, conditioned on state $s$. $\pi(a|s)\in[0,1]$ outputs the probability of taking action $a$ at state $s$. The distribution of trajectories $\tau = (s_1, a_1, \hdots,a_{H-1}, s_H)$ is deterministically dependent on $\pi$ and the MDP, and is defined as
\begin{align}
\rho_{\pi}(\tau) = \rho_0(s_1) \prod_{t=2}^{H} \pi(a_{t-1}|s_{t-1})P_{t-1}(s_t|s_{t-1}, a_{t-1}).\nonumber
\end{align}
%
%
The distribution of the states at time step $t$, induced by running the policy $\pi$ until $t$, is defined  $\forall s_t$:
\begin{align}
d_t^{\pi}(s_{t}) = \sum_{\{s_i,a_i\}_{i\leq t-1}} &\rho_{0}(s_1)\prod_{i=1}^{t-1}\pi(a_i|s_i)P_{i}(s_{i+1}|s_{i},a_{i}). \nonumber
\end{align} Note that the summation above can be replaced by an integral if the state or action space is continuous. The average state distribution $\bar{d}^{\pi}(s) = \sum_{t=1}^H d_t^{\pi}(s) / H$. 

The expected average cost of a policy $\pi$ can be defined with respect to $\rho_{\pi}$ or $\{d_t^{\pi}\}$:
\begin{align}
\mu(\pi) &= \mathop{\mathbb{E}}_{\tau\sim \rho_{\pi}}[\sum_{t=1}^H \bar{c}_t(s_t,a_t)]  = \sum_{t=1}^H\mathop{\mathbb{E}}_{s\sim {d}_t^{\pi}(s),a\sim\pi(a|s)}[\bar{c}_t(s,a)].\nonumber
\end{align}


We define the state-action value $Q_t^{\pi}(s,a)$ (i.e., cost-to-go) for policy $\pi$ at time step $t$ as:
\begin{align}
&Q_t^{\pi}(s_t,a_t) 
&=\bar{c}_t(s_t,a_t) + \mathop{\mathbb{E}}_{s\sim P_t(\cdot|s_t,a_t),a\sim\pi(\cdot|s)}Q^\pi_{t+1}(s,a). \nonumber
\end{align} where the expectation is taken over the randomness of the policy $\pi$ and the MDP. 

We define $\pi^{*}$ as the expert policy (e.g., human demonstrators, search algorithms equipped with ground-truth) and $Q^*_t(s,a)$ as the expert's cost-to-go oracle (note $\pi^*$ may not be optimal, i.e., $\pi^* \not\in \arg\min_{\pi} \mu(\pi)$). Throughout the paper, we assume $Q_t^{*}(s,a)$ is known or can be estimated without bias (e.g., by rolling out $\pi^*$: starting from state $s$, applying action $a$, and then following $\pi^*$ for $H-t$ steps). 

When $\pi$ is represented by a function approximator, we use the notation $\pi_{\theta}$ to represent the policy parametrized by $\theta\in \mathbb{R}^{d}$: $\pi(\cdot|s;\theta)$. In this work we specifically consider optimizing policies in which the parameter dimension $d$ may be large.  We also consider the partially observable setting in our experiments, where the policy $\pi(\cdot|o_1,a_1,...,o_t;\theta)$ is defined over the whole history of partial observations and actions ($o_t$ is generated from the hidden state $s_t$). We use an LSTM-based policy \cite{duan2016benchmarking} where the LSTM's hidden states provide a compressed feature of the history.

\section{Differentiable Imitation Learning}
\label{sec:alg}
Policy based imitation learning aims to learn a policy $\hat \pi$ that approaches the performance of the expert $\pi^*$ in testing time when $\pi^*$ is not available anymore. In order to learn rich policies such as with LSTMs or deep networks~\cite{schulman2015trust}, we derive a \emph{policy gradient} method for imitation learning and sequential prediction. To do this, we leverage the reduction of IL and sequential prediction to online learning as shown in \citep{ross2014reinforcement} to learn policies represented by expressive differentiable function approximators. 

The fundamental idea in \citet{ross2014reinforcement} is to use a no-regret online learner to update policies using the following loss function at each episode $n$:
\begin{align}
\label{eq:general_loss}
\ell_{n}(\pi) = \frac{1}{H}\sum_{t=1}^H \mathop{\mathbb{E}}_{s_t\sim d_t^{\pi_n}}\Big[\mathop{\mathbb{E}}_{a \sim \pi(\cdot | s_t)}[Q_t^*(s_t,a)]\Big].
\end{align}
The loss function intuitively encourages the learner to find a policy that minimize the expert's cost-to-go \emph{under the state distribution resulting from the current learned policy $\pi_n$}.
Specifically, \citet{ross2014reinforcement} suggest an algorithm named \emph{AggreVaTe} (Aggregate Values to Imitate) that uses Follow-the-Leader (FTL)~\cite{shalev2012online} to update policies:$\pi_{n+1} = \arg\min_{\pi\in\Pi}\sum_{i=1}^{n}\ell_n(\pi)$, 
where $\Pi$ is a pre-defined convex policy set.
When $\ell_n(\pi)$ is strongly convex with respect to $\pi$ and $\pi^*\in \Pi$, after $N$ iterations AggreVaTe with FTL can find a policy $\hat{\pi}$:
\begin{align}
\label{eq:AggreVaTe_analysis}
\mu(\hat{\pi}) \leq \mu(\pi^*) - \epsilon_N + {O}(\ln(N)/{N}),
\end{align} where $\epsilon_N =[\sum_{n=1}^N\ell_n(\pi^*) - \min_{\pi}\sum_{n=1}^N\ell_n(\pi)]/N$. Note that $\epsilon_N\geq 0$ and the above inequality indicates that $\hat{\pi}$ can outperform $\pi^*$ when $\pi^*$ is not (locally) optimal (i.e., $\epsilon_n > 0$). Our experimental results support this observation.

A simple implementation of AggreVaTe that aggregates the values (as the name suggests) will require an exact solution to a batch optimization procedure in each episode. 
When $\pi$ is represented by large, non-linear function approximators, the $\arg\min$ procedure generally takes more and more computation time as $n$ increases. 

Online Mirror Descent (OMD) \cite{shalev2012online} are popular for online learning due to its efficiency. Therefore we consider two special cases of  OMD for optimizing sequence of losses $\{\ell_n(\pi)\}_n$: Online Gradient Descent (OGD) \cite{Zinkevich2003_ICML}  and Exponential Gradient Descent (EG) \cite{shalev2012online}, which lead to a regular stochastic policy gradient descent algorithm and a natural policy gradient algorithm, respectively. Also, when applying OGD and EG to $\{\ell_n(\pi)\}_n$, one can show that Eq.~\ref{eq:AggreVaTe_analysis} will hold (with $O(1/\sqrt{N})$), as long as $\ell_n(\pi)$ is convex with respect to $\pi$.

\subsection{Online Gradient Descent}

For discrete actions, the gradient of $\ell_n(\pi_{\theta})$ (Eq.~\ref{eq:general_loss}) with respect to the parameters $\theta$ of the policy can be computed as
\begin{align}
\label{eq:regular_dic_gradient}
\nabla_{\theta}\ell_n(\theta) =\frac{1}{H} \sum_{t=1}^H\mathop{\mathbb{E}}_{s_t\sim d_t^{\pi_{\theta_n}}}\sum_{a}\nabla_{\theta}\pi(a|s_t;\theta) Q_t^*(s_t, a).
\end{align}
For continuous action spaces, we cannot simply replace the summation by integration since in practice it is impossible to evaluate $Q_t^*(s,a)$ for infinitely many $a$, 
so, instead, we use importance weighting to re-formulate $\ell_n$ (Eq.~\ref{eq:general_loss}) as
\begin{align}
\label{eq:importance_weight_continuous}
\ell_n(\pi_{\theta})&=\frac{1}{H}\sum_{t=1}^H\mathop{\mathbb{E}}_{s\sim d_t^{\pi_{\theta_n}},a\sim\pi(\cdot|s;\theta_n)}\frac{\pi(a|s;\theta)}{\pi(a|s;\theta_n)}Q^*_t(s,a) \nonumber \\
&= \frac{1}{H}\mathop{\mathbb{E}}_{\tau\sim\rho_{\pi_{\theta_n}}}\sum_{t=1}^H \frac{\pi(a_t|s_t;\theta)}{\pi(a_t|s_t;\theta_n)}Q^*_t(s_t,a_t).
\end{align} See Appendix~\ref{sec:importance_weight} for the derivation of the above equation.
With this reformulation, the gradient with respect to $\theta$ is
\begin{align}
\label{eq:regular_con_gradient}
\nabla_{\theta}\ell_n(\theta) =\frac{1}{H} \mathop{\mathbb{E}}_{\tau\sim\rho_{\pi_{\theta_n}}}\sum_{t=1}^H \frac{\nabla_{\theta}\pi(a_t|s_t;\theta)}{\pi(a_t|s_t;\theta_n)}Q^*_t(s_t,a_t).
\end{align}
The above gradient computation enables a very efficient update procedure with online gradient descent: $\theta_{n+1} = \theta_n - \eta_n \nabla_{\theta}\ell_n(\theta)|_{\theta = \theta_n}$,
where $\eta_n$ is the learning rate.

\subsection{Policy Updates with Natural Gradient Descent}
We derive a natural gradient update procedure for imitation learning inspired by the success of natural gradient descent in RL~\cite{kakade2002natural,bagnell2003covariant,schulman2015trust}. First, we show that Exponential Gradient Descent (EG) can be leveraged to speed up imitation learning in discrete MDPs. Then we extend EG to continuous MDPs, where we show that, with three steps of approximation, EG leads to a natural gradient update procedure. 

\subsubsection{Exponential Gradient in Discrete MDPs}
For notational simplicity, for each state $s\in\mathcal{S}$, we represent the policy $\pi(\cdot |s)$ as a discrete probability vector $\pi^s \in \Delta(A)$. We also represent $d_t^{\pi}$ as a ${S}$-dimension probability vector from $S$-d simplex, consisting of $d_t^{\pi}(s), \forall s\in\mathcal{S}$. 
For each $s$, we use $Q^*_t(s)$ to denote the $A$-dimension vector consisting of the state-action cost-to-go $Q^*_t(s,a)$ for all $a\in\mathcal{A}$. With this notation, the loss function $\ell_n(\pi)$ from Eq.~\ref{eq:general_loss} can now be written as: $\ell_n(\pi) = \frac{1}{H}\sum_{t=1}^H \sum_{s\in\mathcal{S}} d_t^{\pi_n}(s)( \pi^s\cdot Q^*_t(s))$,
where $a\cdot b$ represents the inner product between vectors $a$ and $b$. Exponential Gradient updates $\pi$ as follows: $\forall s\in\mathcal{S}$,
\begin{align}
\label{eq:eg_argmax}
\pi_{n+1} = \mathop{\arg\min}_{\pi^s\in\Delta(A),\forall s\in\mathcal{S}}&\frac{1}{H}\sum_{t=1}^H \sum_{s\in\mathcal{S}} d_t^{\pi_n}(s)\big( \pi^s\cdot Q^*_t(s)\big) \nonumber \\ &+\sum_{s\in\mathcal{S}}\frac{\bar{d}^{\pi_n}(s)}{\eta_{n,s}}KL(\pi_s \| \pi^s_n), 
\end{align} where $KL(q\|p)$ is the KL-divergence between two probability distributions $q$ and $p$.
This leads to the following closed-form update:
\begin{align}
\label{eq:eg_closed_form}
\pi_{n+1}^s[i] = \frac{\pi_n^s[i]\exp\big(-\eta_{n,s} \tilde{Q}_s^e[i]\big)}{\sum_{j=1}^{|\mathcal{A}|}\pi_n^s[j]\exp\big(-\eta_{n,s} \tilde{Q}_s^e[j]\big)}, i\in [|\mathcal{A}|],
\end{align} where $\tilde{Q}_s^e = \sum_{t=1}^H d_t^{\pi_n}(s)Q^*_t(s) / (H\bar{d}^{\pi_n}(s))$. We refer readers to \cite{shalev2012online} or Appendix~\ref{sec:EG_derivation} for the derivations of the above closed-form updates. 


\subsubsection{Continuous MDPs}
We now consider how to update the parametrized policy $\pi_{\theta}$ for continuous MDPs. Replacing summations by integrals, Eq.~\ref{eq:eg_argmax} can be written as:
\begin{align}
\label{eq:continuous_eg}
\theta = \arg\min_{\theta}\frac{1}{H}& \sum_{t=1}^H  \mathop{\mathbb{E}}_{s\sim d_t^{\pi_{\theta_n}}}\mathop{\mathbb{E}}_{a\sim\pi(\cdot|s;\theta)}[Q_t^*(s,a)] \nonumber\\
& + \mathop{\mathbb{E}}_{s\sim\bar{d}^{\pi_{\theta_n}}}KL(\pi_{\theta} ||\pi_{\theta_n})/\eta_{n}.
\end{align} 
In order to solve for $\theta$ from Eq.~\ref{eq:continuous_eg}, we apply several approximations. We first approximate $\ell_n(\theta)$ (the first part of the RHS of the above equation) by its first-order Taylor expansion: $\ell_n(\theta)\approx \ell_n(\theta_n) + \nabla_{\theta_n}\ell_n(\theta_n) \cdot (\theta - \theta_n)$. When $\theta$ and $\theta_n$ are close, this is a valid local approximation. 

Second, we replace $KL(\pi_{\theta}||\pi_{\theta_n})$ by $KL(\pi_{\theta_n}||\pi_{\theta})$, which is a local approximation since $KL(q||p)$ and $KL(p||q)$ are equal up to the second order \cite{kakade2002approximately,schulman2015trust}.

Third, we approximate $KL(\pi_{\theta_n}||\pi_{\theta})$ by a second-order Taylor expansion around $\theta_n$, such that we can approximate the penalization using the Fisher information matrix:
\begin{align}
\mathop{\mathbb{E}}_{s\sim\bar{d}^{\pi_{\theta_n}}}KL(\pi_{\theta_n} || \pi_{\theta}) \approx (1/2)(\theta - \theta_n)^T I(\theta_n)(\theta-\theta_n), \nonumber
\end{align} where the Fisher information matrix $I(\theta_n)=\mathop{\mathbb{E}}_{s,a\sim \bar{d}^{\pi_{\theta_n}}\pi_{\theta_n}(a|s)}\big(\nabla_{\theta_n}\log(\pi_{\theta_n}(a|s))\big)\big(\nabla_{\theta_n}\log(\pi_{\theta_n}(a|s)\big)^T$. 

Inserting these three approximations into Eq.~\ref{eq:continuous_eg}, and solving for $\theta$, we reach the following update rule $\theta_{n+1} = \theta_n - \eta_{n}I(\theta_n)^{-1}\nabla_{\theta}\ell_n(\theta)|_{\theta =\theta_n}$,
which is similar to the natural gradient update rule developed in \cite{kakade2002natural} 
for the RL setting. 
\citet{bagnell2003covariant} provided an equivalent representation for  Fisher information matrix: 
\begin{align}
\label{eq:fisher_traj}
I(\theta_n )=\frac{1}{H^2}\mathop{\mathop{\mathbb{E}}}_{\tau\sim \rho_{\pi_{\theta_n}}}\nabla_{\theta_n}\log(\rho_{\pi_{\theta_n}}(\tau))\nabla_{\theta_n}\log(\rho_{\pi_{\theta_n}}(\tau))^T,
\end{align} where $\nabla_{\theta}\log(\rho_{\pi_{\tau}}(\tau))$ is the gradient of the log likelihood of the trajectory $\tau$ which can be computed as $\sum_{t=1}^H\nabla_{\theta}\log(\pi_{\theta}(a_t|s_t))$. In the remainder of the paper, we use this Fisher information matrix representation, which yields much faster computation of the descent direction $\delta_\theta$, as we will explain in the next section.

\section{Sample-Based Practical Algorithms}
In the previous section, we derived a regular gradient update procedure and a natural gradient update procedure for IL. Note that all of the computations of gradients and Fisher information matrices assumed it was possible to exactly compute expectations including $\mathop{\mathbb{E}}_{s\sim d^{\pi}}$ and $\mathop{\mathbb{E}}_{a\sim \pi(a|s)}$. In this section, we provide practical algorithms where we approximate the gradients and Fisher information matrices using finite samples collected during policy execution. 

\subsection{Gradient Estimation and Variance Reduction}
We consider an episodic framework where given a policy $\pi_n$ at episode $n$, we roll out $\pi_n$ $K$ times to collect $K$ trajectories $\{\tau_i^{n}\}$, for $i\in [K]$, $\tau_i^n=\{s_1^{i,n},a_1^{i,n},...\}$.  For gradient $\nabla_{\theta}\ell_n(\theta)|_{\theta=\theta_n}$ 
we can compute an unbiased estimate using $\{\tau_i^n\}_{i\in[K]}$:
\begin{align}
\label{eq:gradient_finite_dic}
&\tilde{\nabla}_{\theta_n} = \frac{1}{HK}\sum_{i=1}^K\sum_{t=1}^H\sum_{a}\nabla_{\theta_n}\pi_{\theta_n}(a|s^{i,n}_t) Q_t^*(s^{i,n}_t, a), \\
&\label{eq:gradient_finite_con}
\tilde{\nabla}_{\theta_n} = \frac{1}{HK}\sum_{i=1}^K\sum_{t=1}^H \frac{\nabla_{\theta_n}\pi_{\theta_n}(a^{i,n}_t|s_t^{i,n})}{\pi_{\theta_n}(a_t^{i,n}|s_t^{i,n})}Q_t^*(s_t^{i,n},a_t^{i,n}).
\end{align} for discrete and  continuous setting respectively. 

When we can compute $V_t^*(s)$ (e.g., $\min_a Q_t^*(s,a)$), we can replace $Q_t^*(s_t^{i,n},a)$ in Eq.~\ref{eq:gradient_finite_dic} and Eq.~\ref{eq:gradient_finite_con} by the state-action advantage function $A^*_t(s_t^{i,n},a) = Q^*_t(s_t^{i,n},a) - V_t^*(s_t^{i,n})$, which leads to the following two unbiased and variance-reduced gradient estimations \cite{greensmith2004variance}:
\begin{align}
&\tilde{\nabla}_{\theta_n} = \frac{1}{HK}\sum_{i=1}^K\sum_{t=1}^H\sum_{a}\nabla_{\theta_n}\pi_{\theta_n}(a|s^{i,n}_t) A_t^*(s^{i,n}_t, a), \label{eq:gradient_finite_dic_vr}\\
&\tilde{\nabla}_{\theta_n} = \frac{1}{HK}\sum_{i=1}^K\sum_{t=1}^H \frac{\nabla_{\theta_n}\pi_{\theta_n}(a^{i,n}_t|s_t^{i,n})}{\pi_{\theta_n}(a_t^{i,n}|s_t^{i,n})}A_t^*(s_t^{i,n},a_t^{i,n}),
\label{eq:gradient_finite_con_vr}
\end{align} where Eq.~\ref{eq:gradient_finite_dic_vr} is for discrete action and Eq.~\ref{eq:gradient_finite_con_vr} for continuous action . 

The Fisher information matrix (Eq.~\ref{eq:fisher_traj}) is approximated as:
\begin{align}
\tilde{I}(\theta_n) &= \frac{1}{H^2K}\sum_{i=1}^K \nabla_{\theta_n}\log(\rho_{\pi_{\theta_n}}(\tau_i))\nabla_{\theta_n}\log(\rho_{\pi_{\theta_n}}(\tau_i))^T \nonumber\\
& = S_n S_n^T,
\end{align} where, for notation simplicity, we denote $S_n$ as a $d\times K$ matrix where the $i$'s th column is $\nabla_{\theta_n}\log(\rho_{\pi_{\theta_n}}(\tau_i))/(H\sqrt{K})$. Namely the Fisher information matrix is represented by a sum of $K$ rank-one matrices. For large policies represented by neural networks, $K\ll d$, and hence  $\tilde{I}(\theta_n)$ a low rank matrix. 
One can find the descent direction $\delta_{\theta_n}$ by solving the linear system $S_nS_n^T \delta_{\theta_n} = \tilde{\nabla}_{\theta_n}$ for $\delta_{\theta_n}$ using Conjugate Gradient (CG) with a fixed number of iterations, which is equivalent to solving the above linear systems using the Partial Least Square \cite{phatak2002exploiting}. This approach is used in TRPO \cite{schulman2015trust}. The difference is that our representation of the Fisher matrix is in the form of $S_nS_n^T$ and in CG we never need to explicitly compute or store $S_nS_n^T$ which requires $d^2$ space and time. Instead, we only compute and store $S_n$ ($O(Kd)$) and the total computational time is still $O(K^2 d)$. The learning-rate for natural gradient descent can be chosen as $\eta_n = \sqrt{\delta_{KL}/(\tilde{\nabla}_{\theta_n}^T\delta_{\theta_n})}$, such that $KL(\rho_{\pi_{\theta_{n+1}}(\tau)}\|\rho_{\pi_{\theta_{n}}(\tau)})\approx \delta_{KL}\in\mathbb{R}^+$

\begin{algorithm}[tb]
 \caption{AggreVaTeD (Differentiable AggreVaTe)}
 \begin{algorithmic}[1]
 \label{alg:DIL}
 \STATE {\bfseries Input:} The given MDP and expert $\pi^*$.  Learning rate $\{\eta_n\}$. Schedule rate $\{\alpha_i\}$, $\alpha_n\to0, n\to\infty$. 
 \STATE Initialize policy $\pi_{\theta_1}$ (either random or supervised learning). 
 \FOR{n = 1 to N}
    \STATE Mixing policies: $\hat{\pi}_n= \alpha_n\pi^* + (1-\alpha_n)\pi_{\theta_n}$.
    \STATE \label{line:rolling_in} Starting from $\rho_0$, roll in by executing $\hat{\pi}_{n}$ on the given MDP to generate $K$ trajectories $\{\tau_i^n\}$.
    \STATE Using $Q^*$ and $\{\tau_i^n\}_i$, compute the descent direction $\delta_{\theta_n}$ (Eq.~\ref{eq:gradient_finite_dic}, Eq.~\ref{eq:gradient_finite_con},
    Eq.~\ref{eq:gradient_finite_dic_vr}, Eq.~\ref{eq:gradient_finite_con_vr}, or CG).
    \label{line:gradient_compute}
    \STATE Update: $\theta_{n+1} = \theta_n - \eta_n\delta_{\theta_n}$.
\ENDFOR
 \STATE {\bfseries Return:} the best hypothesis $\hat{\pi}\in\{\pi_n\}_n$ on validation.
 \end{algorithmic}
\end{algorithm}

\subsection{Differentiable Imitation Learning: AggreVaTeD}
We present the differentiable imitation learning framework \emph{AggreVaTeD}, in Alg.~\ref{alg:DIL}. At every iteration $n$, the roll-in policy $\hat{\pi}_n$ is a mix of the expert policy $\pi^*$ and the current policy $\pi_{\theta_n}$, with mixing rate $\alpha$ ($\alpha_n\to0, n\to\infty$): at every step, with probability $\alpha$, $\hat{\pi}_n$ picks $\pi^*$ and else $\pi_{\theta_n}$. This mixing strategy with decay rate was first introduced in \cite{Ross2011_AISTATS} for IL, and later on was used in sequence prediction \cite{bengio2015scheduled}. In Line 6 
one can choose Eq.~\ref{eq:gradient_finite_dic} or the corresponding variance reduced estimation Eq.~\ref{eq:gradient_finite_dic_vr} (Eq.~\ref{eq:gradient_finite_con} and Eq.~\ref{eq:gradient_finite_con_vr} for continuous actions) to perform regular gradient descent, and choose CG to perform natural gradient descent. Compared with previous well-known IL and sequential prediction algorithms \cite{Ross2011_AISTATS,ross2014reinforcement,chang2015learning}, AggreVaTeD is extremely simple: we do not need to perform any Data Aggregation (i.e., we do not need to store all $\{\tau_i\}_i$ from all previous iterations); the computational complexity of each episode scales as $O(d)$. 

When we use non-linear function approximators to represent the polices, the analysis of AggreVaTe from \cite{ross2014reinforcement} will not hold, since the loss function $\ell_n(\theta)$ is not convex with respect to parameters $\theta$. Nevertheless, as we will show in experiments, in practice AggreVaTeD is still able to learn a policy that is competitive with, and sometimes superior to the oracle's performance.

\section{Quantify the Gap: An Analysis of IL vs RL}
\begin{figure}
  \centering
      \includegraphics[trim={1cm 0.5cm 0 1cm},clip, width=0.22\textwidth]{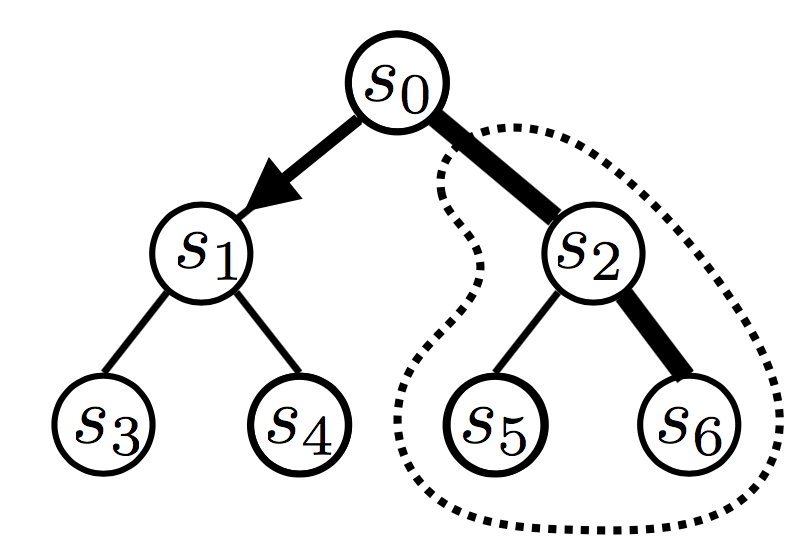}
  \caption{The binary tree structure MDP $\tilde{\mathcal{M}}$. 
  }
  \label{fig:binary_MDP}
  \vspace{-5pt}
\end{figure}
How much faster can IL learn a good policy than RL? In this section we quantify the gap on discrete MDPs when IL can (1) query for an {optimal} $Q^*$ or (2) query for a noisy but unbiased estimate of $Q^*$. To measure the speed of learning, we look at the \emph{cumulative} regret of the entire learning process, defined as $R_N = \sum_{n=1}^N (\mu(\pi_n) - \mu(\pi^*))$. A smaller regret rate indicates faster learning. Throughout this section, we assume the expert $\pi^*$ is optimal. We consider finite-horizon, episodic IL and RL algorithms. 

\subsection{Exponential Gap}
\label{sec:special_mdp}
We consider an MDP $\mathcal{M}$ shown in Fig.~\ref{fig:binary_MDP} which is a depth-K binary tree-structure with $S = 2^K-1$ states and two actions ${a_l, a_r}$:  go-left and go-right. The transition is deterministic and the initial state $s_0$ (root) is fixed. The cost for each non-leaf state is zero; the cost for each leaf is i.i.d sampled from a given distribution (possibly different distributions per leaf).  Below we show that for $\mathcal{M}$, IL can be exponentially more sample efficient than RL.
\begin{theorem}
\label{them:special_lower}
For $\mathcal{M}$, the regret $R_N$ of any finite-horizon, episodic RL algorithm is at least:
\begin{align}
\mathbb{E}[R_N] \geq \Omega(\sqrt{SN}).
\end{align} 
\end{theorem} The expectation is with respect to random generation of cost and internal randomness of the algorithm. However, for the same MDP $\mathcal{M}$, with the access to $Q^*$, we show IL can learn exponentially faster:
\begin{theorem}
\label{them:special_upper}
For the MDP ${\mathcal{M}}$, there exists a policy class such that AggreVaTe with FTL that can achieve the following regret bound:
\begin{align}
R_N \leq O(\ln{(S)}).
\end{align}
\end{theorem}
Fig.~\ref{fig:binary_MDP} illustrates the intuition behind the theorem. Assume during the first episode, the initial policy $\pi_1$ picks the rightmost trajectory (bold black) to explore and the algorithm queries from oracle that for $s_0$ we have $Q^*(s_0,a_l)<Q^*(s_0,a_r)$, it immediately learns that the optimal policy will go left (black arrow) at $s_0$. Hence the algorithm does not have to explore the right sub-tree (dotted circle).

Next we consider a more difficult setting where one can only query for a noisy but unbiased estimate of $Q^*$ (e.g., by rolling out $\pi^*$ finite number of times). The above halving argument will not apply since deterministically eliminating nodes based on noisy estimates  might permanently remove good trajectories. However, IL can still achieve a poly-log regret with respect to $S$, even in the noisy setting:
\begin{theorem}
\label{them:special_lower_noisy}
With only access to unbiased estimate of $Q^*$, for the MDP ${\mathcal{M}}$, AggreVaTeD with EG that can achieve the following regret with probability at least $1-\delta$:
\begin{align}
R_N \leq O\Big(\ln(S)(\sqrt{\ln(S)N}+\sqrt{\ln(2/\delta)N})\Big).
\end{align}
\end{theorem} 
The detailed proofs of the above three theorems can be found in Appendix~\ref{sec:special_lower},\ref{sec:special_upper},\ref{sec:proof_special_noisy} respectively. 
In summary, for MDP ${\mathcal{M}}$, IL is is exponentially faster than RL. 

\subsection{Polynomial Gap and Near-Optimality}
We next quantify the gap in general discrete MDPs and also show that AggreVaTeD is near-optimal. We consider the harder case where we can only access an unbiased estimate of $Q^*_t$, for any $t$ and state-action pair. The policy $\pi$ is represented as a set of probability vectors $\pi^{s,t}\in\Delta(A)$, for all $s \in\mathcal{S}$ and  $t\in[H]$: $\pi = \{\pi^{s,t}\}_{s\in\mathcal{S},t\in[H]}$. 


\begin{theorem}
\label{them:upper_bound}
With access to unbiased estimates of $Q^*_t$, AggreVaTeD with EG achieves the regret upper bound:
\begin{align}
\label{eq:regret_eg}
R_N \leq O\big(HQ_{\max}^e\sqrt{S\ln(A)N}\big).
\end{align}
\end{theorem} Here $Q_{\max}^e$ is the maximum cost-to-go of the expert.\footnote{Here we assume $Q_{\max}^e$ is a constant compared to $H$. If $Q_{\max}^e=\Theta(H)$, then the expert is no better than a random policy of which the cost-to-go is around $\Theta(H)$.}
The total regret shown in Eq.~\ref{eq:regret_eg} allows us to compare IL algorithms to RL algorithms. For example, the Upper Confidence Bound (UCB) based, near-optimal optimistic RL algorithms from \cite{jaksch2010near}, specifically designed for efficient exploration, admit regret $\tilde{O}(HS\sqrt{H AN})$, leading to a gap of approximately $\sqrt{HAS}$ compared to the regret bound of imitation learning shown in Eq.~\ref{eq:regret_eg}. 

We also provide a lower bound on $R_N$ for $H=1$ case which shows the dependencies on $N, A, S$ are tight:
\begin{theorem}
\label{them:lower_bound}
There exists an MDP (H=1), with only access to unbiased estimate of $Q^*$, any finite-horizon episodic imitation learning algorithm  must have:
\begin{align}
\mathbb{E}[R_N] \geq \Omega(\sqrt{S\ln({A})N}).
\end{align}
\end{theorem} The proofs of the above two theorems regarding general MDPs can be found at Appendix~\ref{sec:proof_upper_bound},\ref{sec:proof_lower_bound}. In summary for discrete MDPs, one can expect at least a polynomial gap and a possible exponential gap between IL and RL.


\vspace{-5pt}
\section{Experiments}
\begin{figure*}[t!]
	\centering
	\vspace{-2mm}
	\begin{subfigure}[l]{0.1962\textwidth}
        \includegraphics[width=1.12\textwidth,keepaspectratio]{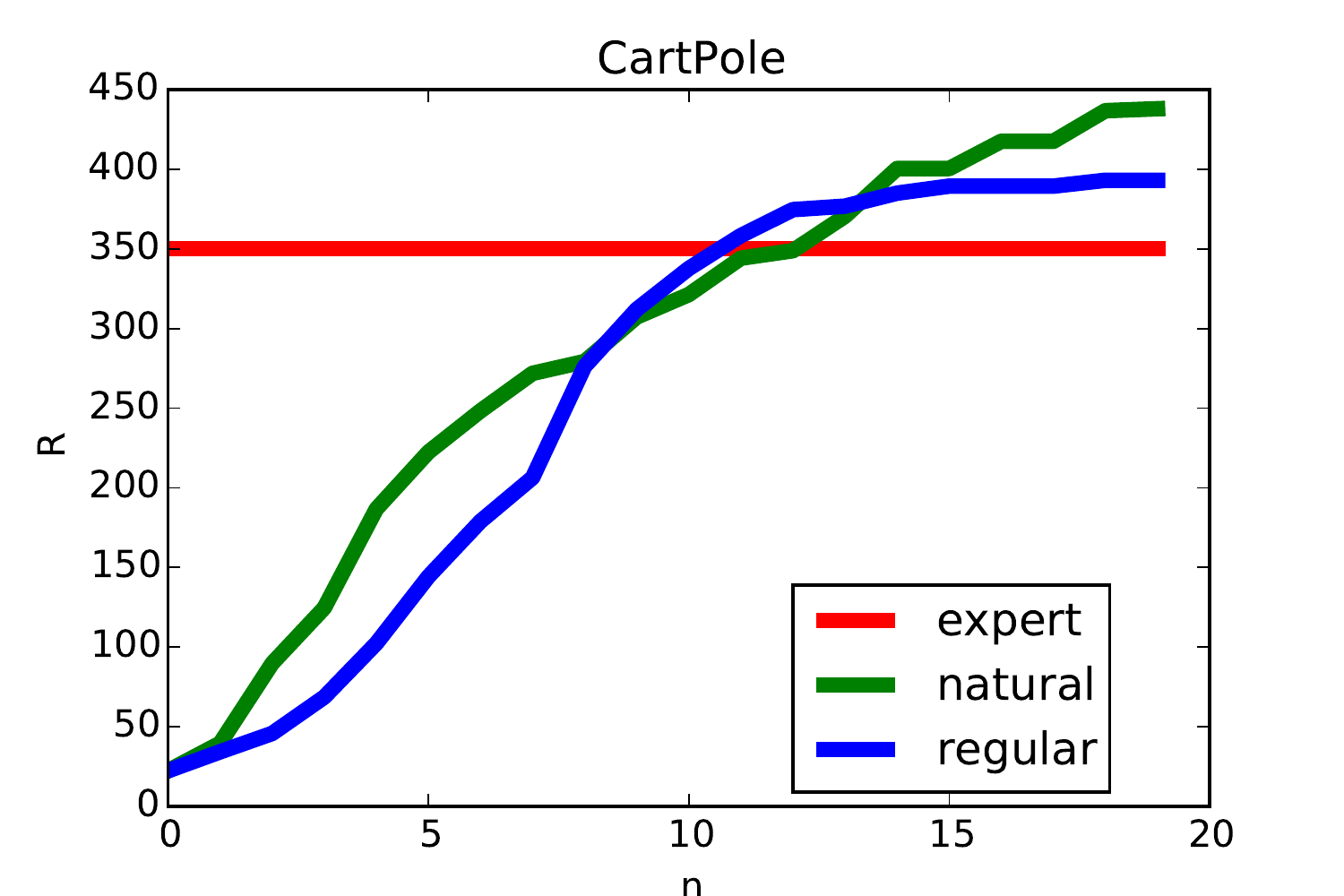}
        \caption{Cartpole}
        \label{fig:cartpole}
    \end{subfigure}
	\begin{subfigure}[l]{0.1962\textwidth}
        \includegraphics[width=1.12\textwidth,keepaspectratio]{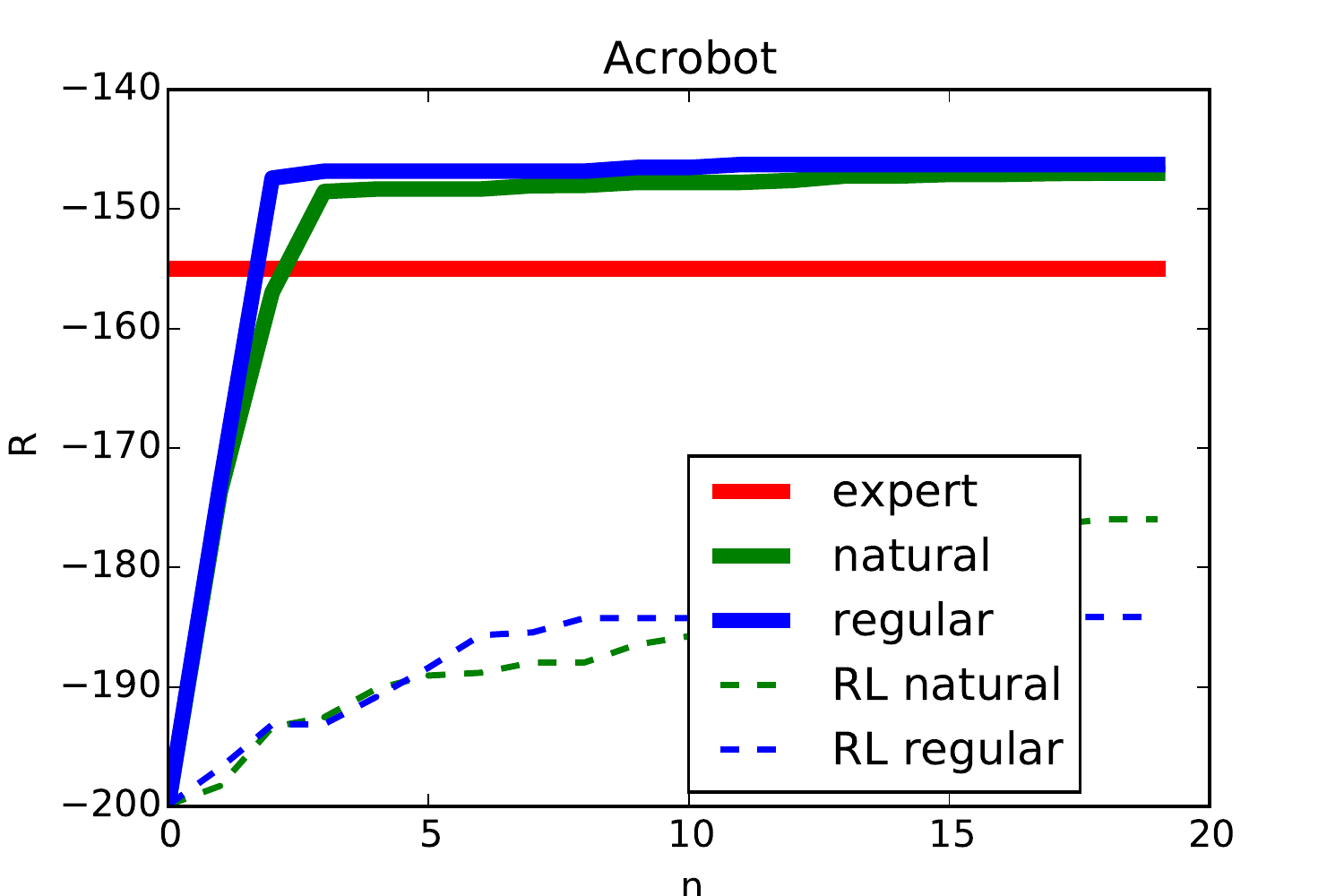}
        \caption{Acrobot}
        \label{fig:acrobot}
    \end{subfigure}
    \begin{subfigure}[l]{0.1962\textwidth}
        \includegraphics[width=1.12\textwidth,keepaspectratio]{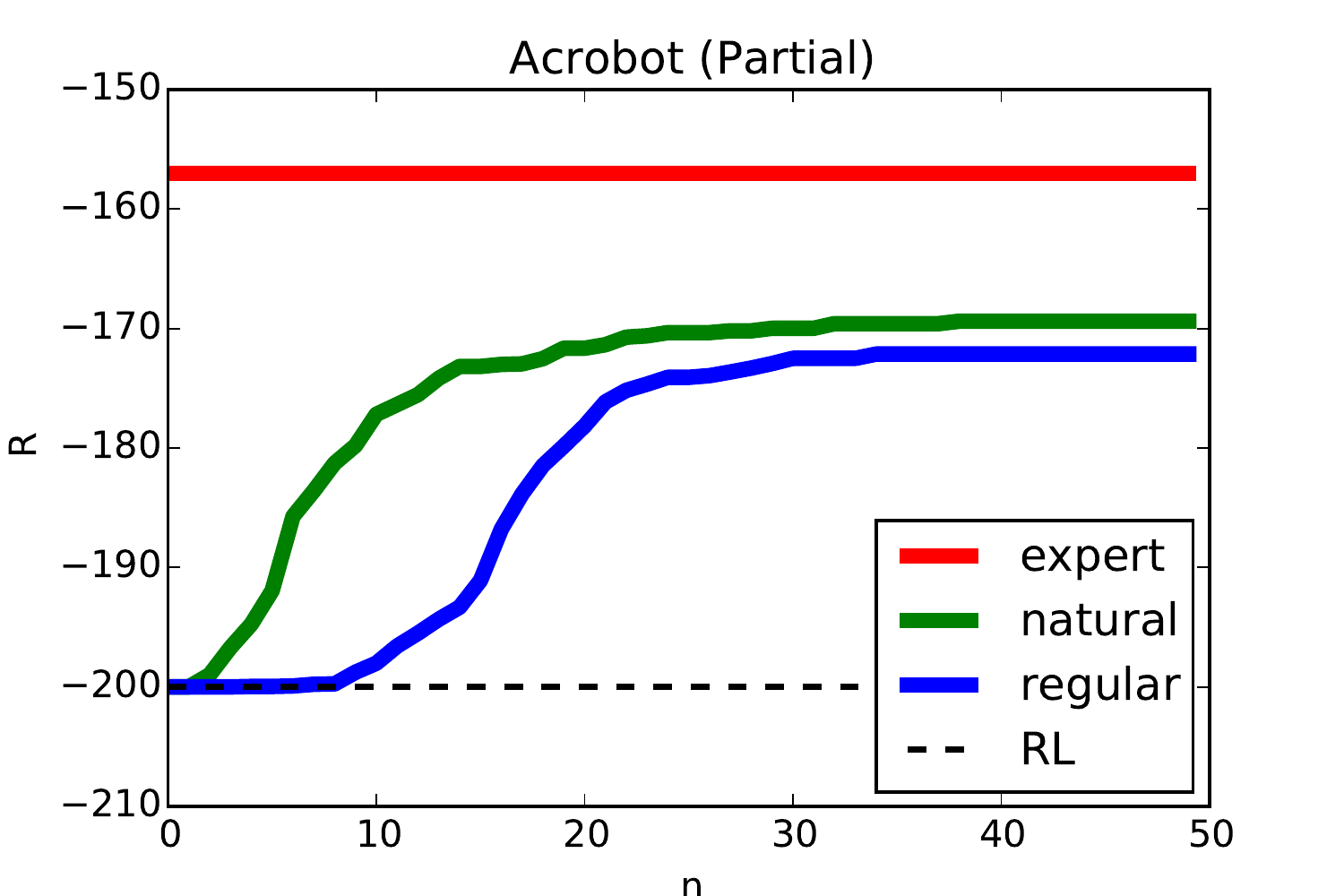}
        \caption{Acrobot (POMDP)}
        \label{fig:acrobot_partial}
    \end{subfigure}
    \begin{subfigure}[l]{0.1962\textwidth}
        \includegraphics[width=1.12\textwidth,keepaspectratio]{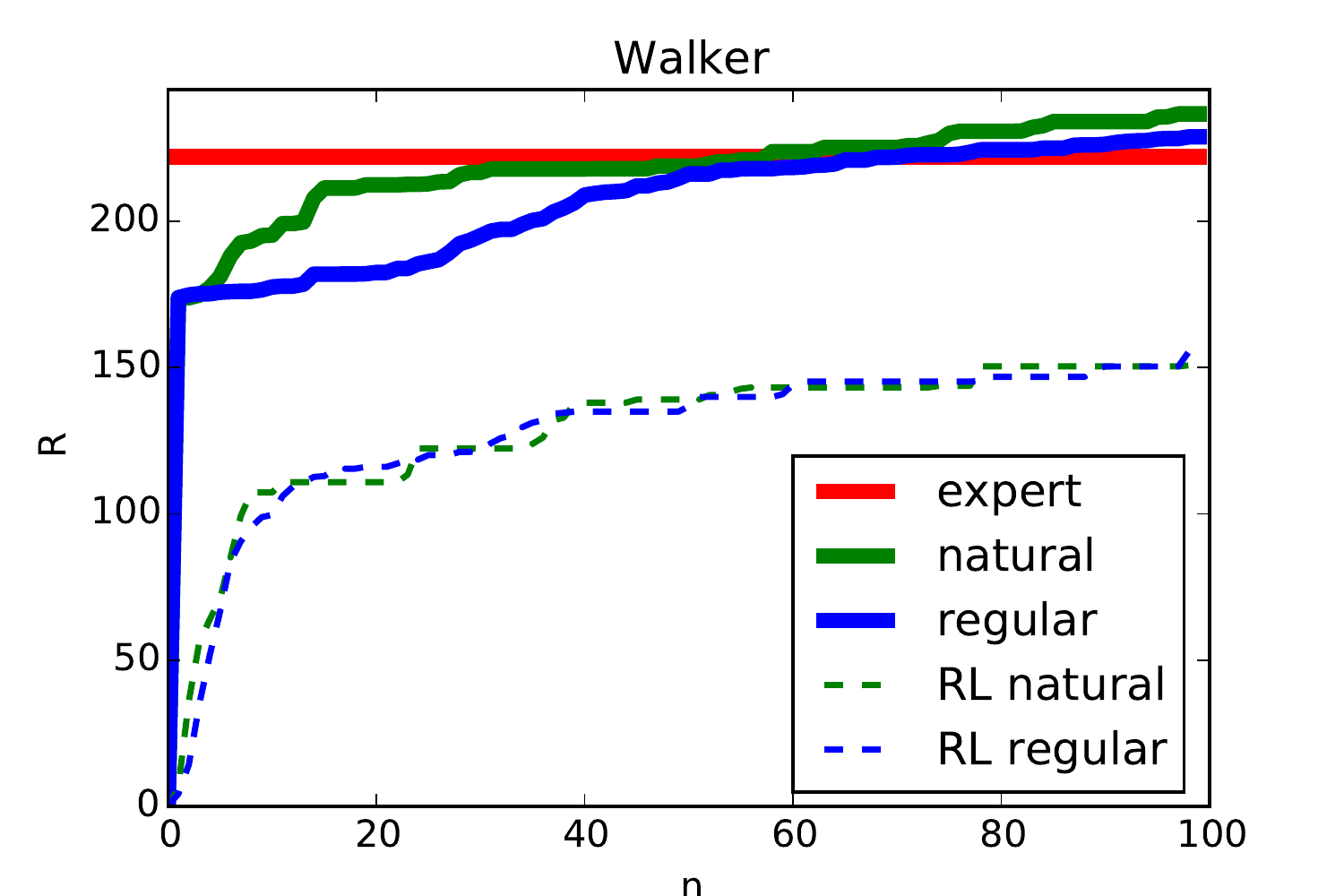}
        \caption{Walker}
        \label{fig:walker}
    \end{subfigure}
	\begin{subfigure}[l]{0.1962\textwidth}
        \includegraphics[width=1.12\textwidth,keepaspectratio]{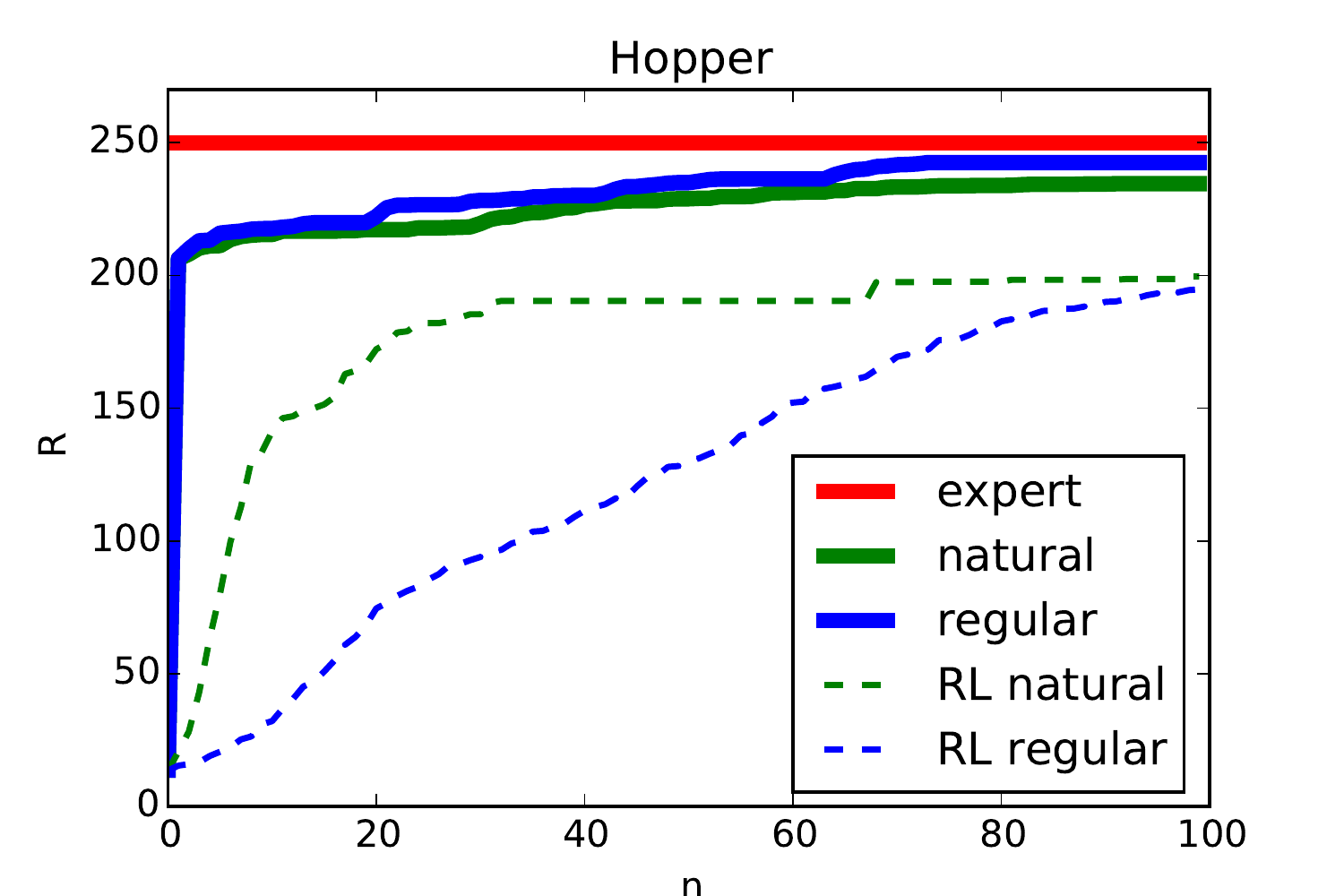}
        \caption{Hopper}
        \label{fig:hopper}
    \end{subfigure}
    \caption{Performance (cumulative reward $R$ on y-axis) versus number of episodes ($n$ on x-axis) of AggreVaTeD (blue and green), experts (red), and RL algorithms (dotted) on different robotics simulators. }
    \label{fig:perf_robotics}
\end{figure*}
We evaluate our algorithms on robotics simulations from OpenAI Gym \cite{brockman2016openai} and on Handwritten Algebra Dependency Parsing \cite{duyckpredicting}. We report reward instead of cost, since OpenAI Gym by default uses reward and dependency parsing aims to maximize UAS score.  As our approach only promises there exists a policy among all of the learned polices that can perform as well as the expert, we report the performance of the best policy so far: $ \max\{\mu(\pi_1), ..., \mu(\pi_i)\}$. For regular gradient descent, we use ADAM \cite{kingma2014adam} which is a first-order no-regret algorithm, and for natural gradient, we use CG to compute the descent direction. For RL we use REINFORCE \cite{williams1992simple} and Truncated Natural Policy Gradient (TNPG) \cite{duan2016benchmarking}.

\subsection{Robotics Simulations}
We consider CartPole Balancing, Acrobot Swing-up, Hopper and Walker. For generating an expert, similar to previous work \cite{ho2016generative}, we used a Deep Q-Network (DQN) to generate $Q^*$ for CartPole and Acrobot (e.g., to simulate the settings where $Q^*$ is available), while using the publicly available TRPO implementation to generate $\pi^*$ for Hopper and Walker to simulate the settings where one has to estimate $Q^*$ by Monte-Carlo roll outs $\pi^*$. 

\paragraph{Discrete Action Setting} We use a one-layer (16 hidden units) neural network with ReLu activation functions to represent the policy $\pi$ for the Cart-pole and Acrobot benchmarks. The value function $Q^*$ is obtained from the DQN \cite{mnih2015human} and represented by a multi-layer fully connected neural network. The policy $\pi_{\theta_1}$ is initialized with common ReLu neural network initialization techniques. For the scheduling rate $\{\alpha_i\}$, we set all $\alpha_i = 0$: namely we did not roll-in using the expert's actions during training.  We set the number of roll outs $K = 50$ and horizon $H = 500$ for CartPole and $H = 200$ for Acrobot.

Fig.~\ref{fig:cartpole} and \ref{fig:acrobot} shows the performance averaged over 10 random trials of AggreVaTeD with regular gradient descent and natural gradient descent. Note that AggreVaTeD outperforms the experts' performance significantly: Natural gradient surpasses the expert by 5.8$\%$ in Acrobot and $\mathbf{25\%}$ in Cart-pole. Also, for Acrobot swing-up, at horizon $H=200$, with high probability a randomly initialized neural network policy won't be able to collect any reward signals. Hence the improvement rates of REINFORCE and TNPG are slow. In fact, we observed that for a short horizon such as $H=200$, REINFORCE and Truncated Natural Gradient often even fail to improve the policy at all (failed 6 times among 10 trials). On the contrary, AggreVaTeD does not suffer from the delayed reward signal issue, since the expert will collect reward signals much faster than a randomly initialized policy. 

Fig.~\ref{fig:acrobot_partial} shows the performance of AggreVaTeD with an LSTM policy (32 hidden states) in a partially observed setting where the expert has access to full states but the learner has access to partial observations (link positions). RL algorithms did not achieve any improvement while AggreVaTeD still achieved 92$\%$ of expert's performance.

\vspace{-5pt}
\paragraph{Continuous Action Setting}
We test our approaches on two robotics simulators with continuous actions: (1) the 2-d Walker and (2) the Hopper from the MuJoCo physics simulator. Following the neural network settings described in \citet{schulman2015trust}, the expert policy $\pi^*$ is obtained from TRPO with one hidden layer (64 hidden states), which is the same structure that we use to represent our policies $\pi_{\theta}$. We set $K = 50$ and $H = 100$. We initialize $\pi_{\theta_1}$ by collecting $K$ expert demonstrations and then maximize the likelihood of these demonstrations (i.e., supervised learning). We use Eq.~\ref{eq:gradient_finite_con} instead of the variance reduced equation here since we need to use MC roll-outs to estimate $V^*$ (we simply use one roll-out to estimate $Q^*$). 

Fig.~\ref{fig:walker} and \ref{fig:hopper} show the performance averaged over 5 random trials. Note that AggreVaTeD outperforms the expert in the Walker by 5.4$\%$ while achieving 97$\%$ of the expert's performance in the Hopper problem. After 100 iterations, we see that by leveraging the help from experts, AggreVaTeD can achieve much faster improvement rate than the corresponding RL algorithms.

\subsection{Dependency Parsing on Handwritten Algebra}
\begin{table*}[t!]
\begin{center}
\resizebox{1.\textwidth}{!}{ 
    \begin{tabular}{llllllllllr}
   \toprule
    Arc-Eager & AggreVaTeD (LSTMs)   & AggreVaTeD (NN)   & SL-RL (LSTMs) & SL-RL(NN) & RL (LSTMs) & RL (NN) & DAgger & SL (LSTMs) & SL (NN) & Random \\ 
    \midrule
    {Regular} & \textbf{0.924}$\pm$0.10 & 0.851$\pm$0.10 & 0.826$\pm$ 0.09& 0.386$\pm$0.1 & 0.256$\pm$0.07 & 0.227$\pm$0.06 & \multirow{2}{*}{0.832$\pm$0.02} & \multirow{2}{*}{0.813$\pm$0.1} 
    & \multirow{2}{*}{0.325$\pm$0.2}
    &\multirow{2}{*}{$\sim$0.150}\\
    
    {Natural} & 0.915$\pm$0.10 & 0.800$\pm$0.10 & 0.824$\pm$0.10 & 0.345$\pm$0.1 & 0.237$\pm$0.07 &0.241$\pm$0.07 \\
    \bottomrule
    \end{tabular}}
\end{center}
\vspace{-5pt}
\caption{ Performance (UAS) of different approaches on handwritten algebra dependency parsing. \emph{SL} stands for supervised learning using expert's samples: maximizing the likelihood of expert's actions under the sequences generated by expert itself. \emph{SL-RL} means RL with initialization using SL. \emph{Random} stands for the initial performances of random policies (LSTMs and NN).
The performance of DAgger with Kernel SVM is from \cite{duyckpredicting}.\vspace{-2mm}} 
\label{tab:eager}
\end{table*}

We consider a sequential prediction problem: transition-based dependency parsing for handwritten algebra with raw image data \cite{duyckpredicting}. The parsing task for algebra is similar to the classic dependency parsing for natural language \cite{chang2015learning_dependency} where the problem is modelled in the IL setting and the state-of-the-art is achieved by AggreVaTe with FTRL (using Data Aggregation). The additional challenge here is that the inputs are handwritten algebra symbols in raw images. We directly learn to predict parse trees from low level image features (Histogram of Gradient features (HoG)). During training, the expert is constructed using the ground-truth dependencies in training data. 
The full state $s$ during parsing consists of three data structures: Stack, Buffer and Arcs, which store raw images of the algebraic symbols. Since the sizes of stack, buffer and arcs change during parsing, a common approach is to featurize the state $s$ by taking the features of the latest three symbols from stack, buffer and arcs (e.g., \cite{chang2015learning_dependency}). Hence the problem falls into the \emph{partially observable} setting, where the feature $o$ is extracted from state $s$ and only contains partial information about $s$. 
The dataset consists of 400 sets of handwritten algebra equations. We use 80$\%$ for training, 10$\%$ for validation, and 10$\%$ for testing. We include an example of handwritten algebra equations and its dependency tree in Appendix~\ref{sec:parsing_example}. Note that different from robotics simulators where at every episode one can get fresh data from the simulators, the dataset is fixed and sample efficiency is critical. 

\begin{figure}[t!]
	\centering
	\vspace{-2mm}
	\begin{subfigure}[l]{0.238\textwidth}
        \includegraphics[width=1.11\textwidth,keepaspectratio]{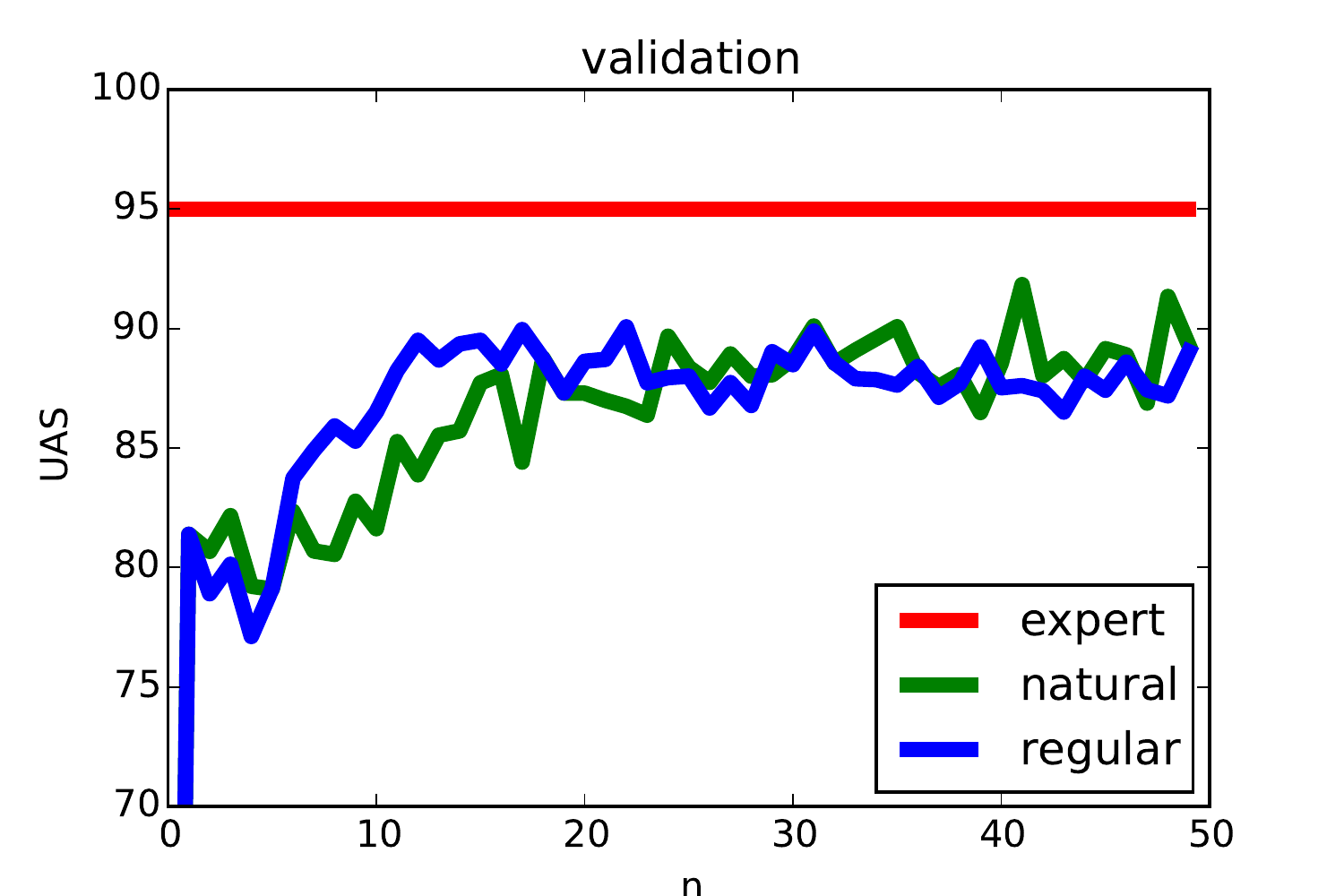}
        \caption{Validation}
        \label{fig:cartpole}
    \end{subfigure}
	\begin{subfigure}[l]{0.238\textwidth}
        \includegraphics[width=1.11\textwidth,keepaspectratio]{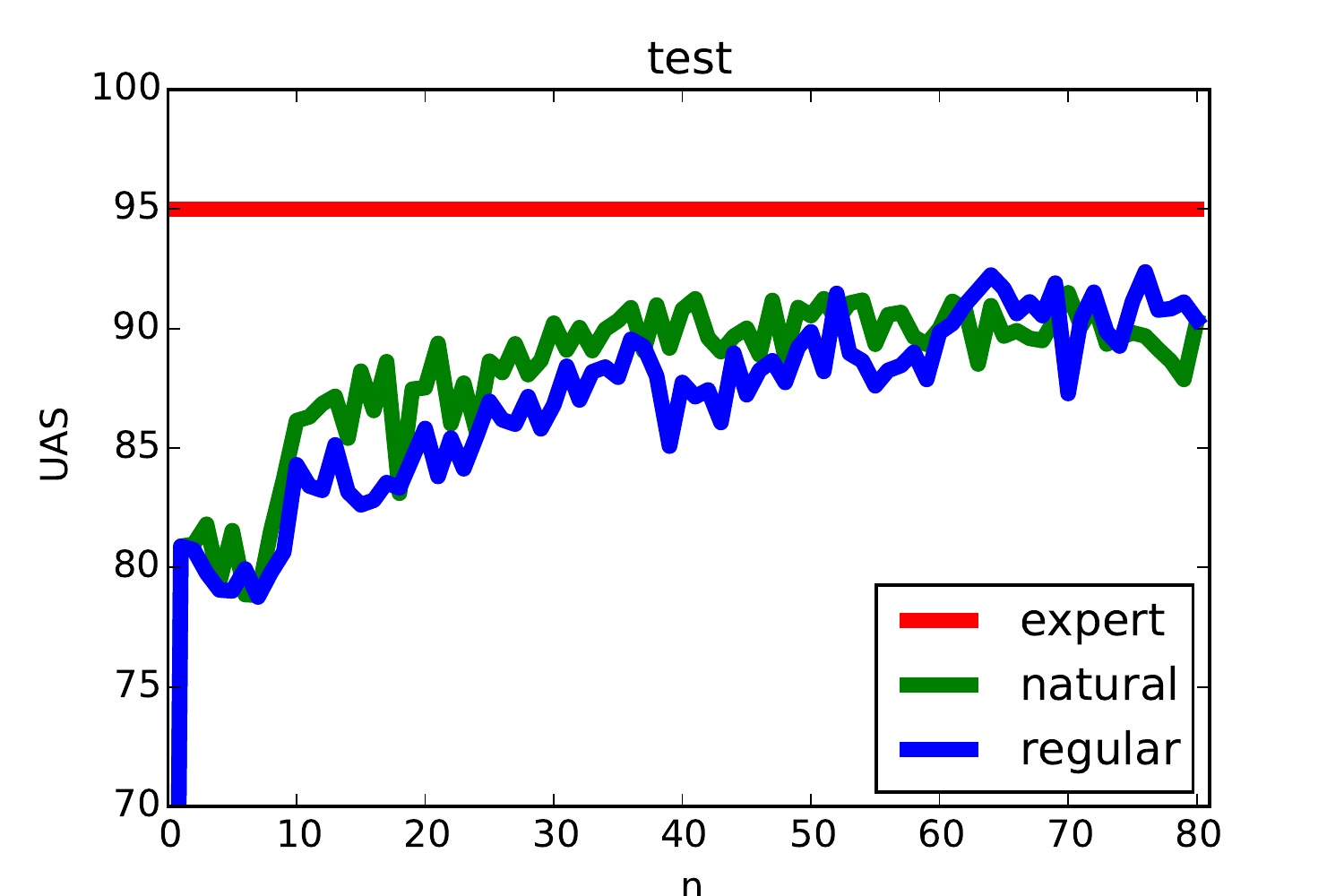}
        \caption{Test}
        \label{fig:acrobot}
    \end{subfigure}
    \caption{UAS (y-axis) versus number of iterations ($n$ on x-axis) of AggreVaTeD with LSTM policy (blue and green), experts (red) on validation set and test set for Arc-Eager Parsing. }
    \label{fig:perf_eager}
    \vspace{-0.25in}
\end{figure}

The RNN policy follows the design from \cite{sutskever2014sequence}. It consists of two LSTMs. Given a sequence of algebra symbols $\tau$, the first LSTM processes one symbol at a time and at the end outputs its hidden states and memory (i.e., a summary of $\tau$). The second LSTM initializes its own hidden states and memory using the outputs of the first LSTM. At every parsing step $t$, the second LSTM takes the current partial observation $o_t$ ($o_t$ consists of features of the most recent item from stack, buffer and arcs) as input, and uses its internal hidden state and memory to compute the action distribution $\pi(\cdot|o_1,...,o_t,\tau)$ conditioned on history. We also tested reactive policies constructed as fully connected ReLu neural networks (NN) (one-layer with 1000 hidden states) that directly maps from observation $o_t$ to action $a$, where $o_t$ uses the most three recent items. We use variance reduced gradient estimations, which give better performance in practice. The performance is summarised in Table~\ref{tab:eager}. Due to the partial observability of the problem, AggreVaTeD with a LSTM policy achieves significantly better UAS scores compared to the NN reactive policy and DAgger with a Kernelized SVM \cite{duyckpredicting}. Also AggreVaTeD with a LSTM policy achieves 97$\%$ of optimal expert's performance. Fig.~\ref{fig:perf_eager} shows the improvement rate of regular gradient and natural gradient on both validation set and test set. Overall we observe that both methods have similar performance. Natural gradient achieves a better UAS score in validation and converges slightly faster on the test set but also achieves a lower UAS score on test set.


\vspace{-5pt}
\section{Conclusion}
We introduced AggreVaTeD, a differentiable imitation learning algorithm which trains neural network policies for sequential prediction tasks such as continuous robot control and dependency parsing on raw image data. We showed that in theory and in practice IL can learn much faster than RL with access to optimal cost-to-go oracles. The IL learned policies were able to achieve expert and sometimes super-expert levels of performance in both fully observable and partially observable settings. The theoretical and experimental results suggest that IL is significantly more effective than RL for sequential prediction with near optimal cost-to-go oracles.

\vspace{-5pt}


{\small
\bibliography{reference}
\bibliographystyle{icml2016}
}

\newpage
\onecolumn
\appendix
\paragraph{Appendix: Proofs and Detailed Bounds}

\section{Derivation of Eq.~\ref{eq:importance_weight_continuous}}
\label{sec:importance_weight}
Starting from Eq.~\ref{eq:general_loss} with parametrized policy $\pi_{\theta}$, we have:
\begin{align}
&\ell_n(\theta) = \frac{1}{H}\sum_{t=1}^H\mathop\mathbb{E}_{s_t\sim d_t^{\pi_{\theta_n}}}\big[ \mathop\mathbb{E}_{a_t\sim\pi(\cdot|s_t;\theta)}[Q_t^*(s_t,a_t)]  \big] \nonumber\\
& = \frac{1}{H}\sum_{t=1}^H\mathop\mathbb{E}_{s_t\sim d_t^{\pi_{\theta_n}}}\big[\int_{a}\pi(a|s_t;\theta)Q_t^*(s_t,a)da  \big] \nonumber\\
& = \frac{1}{H}\sum_{t=1}^H\mathop\mathbb{E}_{s_t\sim d_t^{\pi_{\theta_n}}}\big[\int_{a}\pi(a|s_t;\theta_n)\frac{\pi(a|s_t;\theta)}{\pi(a|s_t;\theta_n)}Q_t^*(s_t,a) da\big] \nonumber\\
& = \frac{1}{H}\sum_{t=1}^H\mathop\mathbb{E}_{s_t\sim d_t^{\pi_{\theta_n}}}\big[\mathop\mathbb{E}_{a\sim\pi(\cdot|s_t;\theta_n)}  \frac{\pi(a|s_t;\theta)}{\pi(a|s_t;\theta_n)}Q_t^*(s_t,a)\big] \nonumber\\
& = \frac{1}{H}\sum_{t=1}^H \mathop\mathbb{E}_{s_t\sim d_t^{\pi_{\theta_n}},a_t\sim \pi(a|s_t;\theta_n)} \Big[\frac{\pi(a_t|s_t;\theta)}{\pi(a_t|s_t;\theta_n)}Q_t^*(s_t,a_t)\Big].
\end{align}

\section{Derivation of Exponential Gradient Update in Discrete MDP}
\label{sec:EG_derivation}
We show the detailed derivation of Eq.~\ref{eq:eg_closed_form} for AggreVaTeD with EG in discrete MDP. Recall that with $KL$-divergence as the penalization, one update the policy in each episode as:
\begin{align}
\{\pi_{n+1}^s\}_{s\in\mathcal{S}} &= \arg\min_{\{\pi^s\in \Delta(A),\forall s\}}\frac{1}{H}\sum_{t=1}^H\sum_{s\in\mathcal{S}}d_t^{\pi_n}(s)\big( \pi^s\cdot Q_t^*(s)\big) + \sum_{s\sim\mathcal{S}}\frac{\bar{d}^{\pi_n}(s)}{\eta_{n,s}}KL(\pi_s \| \pi_n^s) \nonumber
\end{align} Note that in the above equation, for a particular state $s$, optimizing $\pi^s$ is in fact independent of $\pi^{s'}, \forall s'\neq s$. Hence the optimal sequence $\{\pi^s\}_{s\in\mathcal{S}}$ can be achieved by optimizing $\pi^s$ independently for each $s\in\mathcal{S}$. For $\pi^s$, we have the following update rule:
\begin{align}
\pi_{n+1}^s = &\arg\min_{\pi^s\in\Delta(A)}\frac{1}{H}\sum_{t=1}^H d_t^{\pi_n}(s)(\pi^s\cdot Q^*_t(s)) + \frac{\bar{d}^{\pi_n}(s)}{\eta_{n,s}}KL(\pi_s\| \pi_n^s) \nonumber\\
& = \arg\min_{\pi^s\in\Delta(A)}\pi^{s}\cdot (\sum_{t=1}^H d_t^{\pi_n}(s)Q_t^*(s) / H) + \frac{\bar{d}^{\pi_n}(s)}{\eta_{n,s}}KL(\pi_s\|\pi_n^s) \nonumber\\
& = \arg\min_{\pi^s\in\Delta(A)}\pi^{s}\cdot (\sum_{t=1}^H d_t^{\pi_n}(s)Q_t^*(s) /(H\bar{d}^{\pi_n}(s))) + \frac{1}{\eta_{n,s}}KL(\pi_s\|\pi_n^s) \nonumber\\
& = \arg\min_{\pi^s\in\Delta(A)}\pi^s\cdot \tilde{Q}^e(s) + \frac{1}{\eta_{n,s}}\sum_{j=1}^A \pi^s[j](\log(\pi^s[j]) - \log(\pi_n^s[j]))
\end{align} Take the derivative with respect to $\pi^s[j]$, and set  it to zero, we get:
\begin{align}
\tilde{Q}^e(s)[j] +\frac{1}{\eta_{n,s}}(\log(\pi^s[j]/\pi_n^s[j]) + 1) = 0,
\end{align} this gives us:
\begin{align}
\pi^s[j] = \pi_n^s[j]\exp(-\eta_{n,s}\tilde{Q}^e(s)[j]-1).
\end{align} Since $\pi^s\in\Delta(A)$, after normalization, we get:
\begin{align}
\pi^s[j] = \frac{\pi_n^s[j]\exp(-\eta_{n,s}\tilde{Q}^e(s)[j])}{\sum_{i=1}^A \pi_n^s[i]\exp(-\eta_{n,s}\tilde{Q}^e(s)[i])}
\end{align}

\section{Lemmas}
Before proving the theorems, we first present the \emph{Performance Difference Lemma} \cite{kakade2002approximately,ross2014reinforcement} which will be used later:
\begin{lemma}
\label{lemma:performance_difference}
For any two policies $\pi_1$ and $\pi_2$, we have:
\begin{align}
\mu(\pi_1) - \mu(\pi_2) = H\sum_{t=1}^H \mathbb{E}_{s_t\sim d_t^{\pi_1}}\big[\mathbb{E}_{a_t\sim \pi_1(\cdot|s_t)}[Q_t^{\pi_2}(s_t,a_t) - V_t^{\pi_2}(s_t)]\big].
\end{align}
\end{lemma} We refer readers to \cite{ross2014reinforcement} for the detailed proof of the above lemma. 

The second known result we will use is the analysis of Weighted Majority Algorithm. Let us define the linear loss function as $\ell_n(w) = w\cdot y_n$, for any $y_n\in\mathbb{R}^d$, and $w\in\Delta(d)$ from a probability simplex. Running Exponential Gradient Algorithm on the sequence of losses $\{w\cdot y_n\}$ to compute a sequence of decisions $\{w_n\}$, we have:
\begin{lemma} 
\label{lemma:EG}
The sequence of decisions $\{w_n\}$ computed by running Exponential Gradient descent with step size $\mu$ on the loss functions $\{w\cdot y_n\}$ has the following regret bound:
\begin{align}
\sum_{n=1}^N w_n\cdot y_n - \min_{w^*\in\Delta(d)} \sum_{n=1}^N w^*\cdot y_n \leq \frac{\ln(d)}{\mu} + \frac{\mu}{2}\sum_{n=1}^N\sum_{i=1}^d w_n[i] y_n[i]^2.
\end{align}
\end{lemma} We refer readers to \cite{shalev2012online} for detailed proof.

\section{Proof of Theorem~\ref{them:special_lower}}
\label{sec:special_lower}
\begin{proof}
We construct a reduction from stochastic Multi-Arm Bandits (MAB) to the MDP $\tilde{\mathcal{M}}$. A stochastic MAB is defined by $S$ arms denoted as $I^1, ..., I^S$. Each arm $I^t$'s cost $c_{i}$ at any time step $t$ is sampled from a fixed but unknown distribution. A bandit algorithm picks an arm $I_t$ at  iteration $t$ and then receives an unbiased sample of the picked arm's cost $c_{I_t}$. For any bandit algorithm that picks arms $I_1, I_2,...,I_N$ in $N$ rounds, the expected regret is defined as:
\begin{align}
\mathbb{E}[R_N] = \mathbb{E}[  \sum_{n=1}^N c_{I_n}] - \min_{i\in[S]}\sum_{n=1}^N \bar{c}_{i},
\end{align} where the expectation is taken with respect to the randomness of the cost sampling process and possibly the randomness of the bandit algorithm. It has been shown that there exists a set of distributions  from which the arms' costs sampled from, the expected regret $\mathbb{E}[R_N]$ is at least $\Omega(\sqrt{S N})$ \cite{bubeck2012regret}. 

Consider a MAB with $2^{K}$ arms. To construct a MDP from a MAP, we construct a $K+1$-depth binary-tree structure MDP with $2^{K+1}-1$ nodes. We set each node in the binary tree as a state in the MDP. The number of actions of the MDP is two, which corresponds to go left or right at a node in the binary tree. We associate each leaf nodes with arms in the original MAB: the cost of the $i$'th leaf node is sampled from the cost distribution for the $i$'th arm, while the non-leaf nodes have cost always equal to zero. The initial distribution $\rho_0$ concentrates on the root of the binary tree. Note that there are total $2^K$ trajectories from the root to leafs, and we denote them as $\tau_1,...\tau_{2^K}$. We consider finite horizon ($H=K+1$) episodic RL algorithms that outputs $\pi_1,\pi_2,...,\pi_N$ at $N$ episodes, where $\pi_n$ is any deterministic policy that maps a node to actions \emph{left} or \emph{right}. Any RL algorithm must have the following regret lower bound:
\begin{align}
\label{eq:bandit_to_RL}
\mathbb{E}[\sum_{n=1}^N \mu(\pi_n)] - \min_{\pi^*}\sum_{n=1}^N \mu(\pi^*) \geq \Omega(\sqrt{SN}),
\end{align} where the expectation is taken with respect to the possible randomness of the RL algorithms. Note that any deterministic policy $\pi$ identifies a trajectory in the binary tree when rolling in from the root. The optimal policy $\pi^*$ simply corresponds to the trajectory that leads to the leaf with the mininum expected cost. Note that each trajectory is associated with an arm from the original MAB, and the expected total cost of a trajectory corresponds to the expected cost of the associated arm. Hence if there exists an RL algorithm that achieves regret $O(\sqrt{SN})$, then we can solve the original MAB problem by simply running the RL algorithm on the constructed MDP. Since the lower bound for MAB is $\Omega(\sqrt{SN})$, this concludes that Eq.~\ref{eq:bandit_to_RL} holds. 
\end{proof}

\section{Proof of Theorem~\ref{them:special_upper}}
\label{sec:special_upper}
\begin{proof}
For notation simplicity we denote $a_l$ as the go-left action while $a_r$ is the go-right action. Without loss of generality, we assume that the leftmost trajectory has the lowest total cost (e.g., $s_3$ in Fig.~\ref{fig:binary_MDP} has the lowest average cost).
We consider the deterministic policy class $\Pi$ that contains all policy $\pi: \mathcal{S}\to \{a_l,a_r\}$. Since there are $S$ states and 2 actions, the total number of policies in the policy class is $2^S$. To prove the upper bound $R_N\leq O(\log(S))$, we claim that for any $e\leq K$, at the end of episode $e$, AggreVaTe with FTL identifies the $e$'th state on the best trajectory, i,e, the leftmost trajectory $s_0, s_1, s_3, ..., s_{(2^{K-1}-1)}$. We can prove the claim by induction. 

At episode $e=1$, based on the initial policy, AggreVaTe picks a trajectory $\tau_1$ to explore. 
AggreVaTe with FTL collects the states $s$ at $\tau_1$ and their associated cost-to-go vectors $[Q^*(s,a_l), Q^*(s,a_r)]$. Let us denote $D_1$ as the dataset that contains the state,cost-to-go pairs: $D_1 = \{(s, [Q^*(s,a_l),Q^*(s,a_l)])\}$, for $s\in \tau_1$. Since $s_0$ is visited, the state-cost pair $(s_0, [Q^*(s_0,a_l),Q^*(s_0,a_r)])$ must be in $D_1$. To update policy from $\pi_1$ to $\pi_2$, AggreVaTe with FTL runs cost-sensitive classification $D_1$ as:
\begin{align}
\label{eq:cs}
\pi_2 = \arg\min_{\pi}\sum_{k=1}^{|D_1|} Q^*(s_k, \pi(s_k)),
\end{align} where $s_k$ stands for the $k$'th data point collected at dataset $D_1$. Due to the construction of policy class $\Pi$, we see that $\pi_2$ must picks action $a_l$ at state $s_0$ since $Q(s_0,a_l)<Q(s_0,a_r)$. Hence at the end of the episode $e=1$, $\pi_2$ identifies $s_1$ (i.e., running $\pi_2$ from root $s_0$ leads to $s_1$), which is on the optimal trajectory.  

Now assume that at the end of episode $n-1$, the newly updated policy $\pi_{n}$ identifies the state $s_{(2^{n-1}-1)}$: namely at the beginning of episode $n$, if we roll-in $\pi_n$, the algorithm will keep traverse along the leftmost trajectory till at least state $s_{(2^{n-1}-1)}$. At episode $n$, let $D_n$ as the dataset contains all data points from $D_{n-1}$ and the new collected state, cost-to-go pairs from $\tau_n$: $D_n = D_{n-1}\cup \{(s, [Q^*(s,a_l),Q^*(s,a_r)])\} $, for all $s\in\tau_n$. Now if we compute policy $\pi_{n+1}$ using cost-sensitive classification (Eq.~\ref{eq:cs}) over $D_n$, we must learn a policy $\pi_{n+1}$ that identifies action $a_l$ at state $s_{(2^{j}-1)}$, since $Q^{e}(s_{(2^{j}-1)}, a_l)< Q^*(s_{(2^{j}-1)}, a_r)$, and $s_{(2^j - 1)}$ is included in $D_n$, for $j=1,..., n-1$.  Hence at the end of episode $n$, we identify a policy $\pi_{n+1}$ such that if we roll in policy $\pi_{n+1}$ from $s_0$, we will traverse along the left most trajectory till we reach $s_{(2^n-1)}$.  

Hence by the induction hypothesis, at the end of episode $K-1$, $\pi_{K}$ will reach state $s_{(2^{K-1}-1)}$, the end of the best trajectory.

Since AggreVaTe with FTL with policy class $\Pi$ identifies the best trajectory with at most $K-1$ episodes, the cumulative regret is then at most $O(K)$, which is $O(\log(S))$ (assuming the average cost at each leaf is a bounded constant), as $S$ is the number of nodes in the binary-tree structure MDP $\tilde{\mathcal{M}}$.
\end{proof}

\section{Proof of Theorem~\ref{them:special_lower_noisy}}
\label{sec:proof_special_noisy}
Since in Theorem~\ref{them:special_lower_noisy} we assume that we only have access to the noisy, but unbiased estimate of $Q^*$, the problem becomes more difficult since unlike in the proof of Theorem~\ref{them:special_upper}, we cannot simply eliminate states completely since the cost-to-go of the states queried from expert is noisy and completely eliminate nodes will potentially result elimination of low cost trajectories. Hence here we consider a different policy representation. We define $2^{K}$ deterministic base policies $\pi^1, ..., \pi^{2^K}$, such that rolling in policy $\pi^i$ at state $s_0$ will traverse along the trajectory ending at the $i$'th leaf. We define the policy class $\Pi$ as the convex hull of the base policies $\Pi = \{\pi: \sum_{i=1}^{2^K}w_i\pi^i, \sum_i^{2^K}w_i = 1, w_i\geq 0, \forall i\}$. Namely each $\pi\in\Pi$ is a stochastic policy: when rolling in, with probability $w_i$, $\pi$ execute the $i$'th base policy $\pi^i$ from $s_0$. Below we prove that AggreVaTeD with Exponential Gradient Descent achieves the regret bound $O(\sqrt{\ln(S) N})$.

\begin{proof}
We consider finite horizon, episodic imitation learning setting where at each episode $n$, the algorithm can roll in the current policy $\pi_n$ once and only once and traverses through trajectory $\tau_n$ . Let us define $\tilde{\ell}_n(w) = \frac{1}{K+1} \sum_{s\in\tau_n}\sum_{j=1}^{2^K}w_j \tilde{Q}^e(s,\pi^j(s))$, where $\tau_n$ is the trajectory traversed by rolling in policy $\pi_n$ starting at $s_0$, and $\tilde{Q}^e$ is a noisy but unbiased estimate of $Q^*$. We simply consider the setting where $\tilde{Q}^e$ is bounded $|\tilde{Q}^e| \leq l_{\max}$ (note that we can easily extend our analysis to a more general case where $\tilde{Q}^e$ is from a sub-Gaussian distribution). Note that $\tilde{\ell}_n(w)$ is simply a linear loss with respect to $w$:
\begin{align}
\tilde{\ell}_n(w) = w\cdot q_n,
\end{align} where $q_n[j] = \sum_{s\in\tau_n} \tilde{Q}^e(s,\pi^j(s))/(K+1)$. AggreVaTeD with EG updates $w$ using Exponential gradient descent. Using the result from lemma~\ref{lemma:EG}, we get:
\begin{align}
&\sum_{n=1}^N (\tilde{\ell}_n(w_n) - \tilde{\ell}_n(w^*)) = \sum_{n=1}^N (w_n \cdot q_n - w^*\cdot q_n) \leq \frac{\ln(2^K)}{\mu} + \frac{\mu}{2}\sum_{n=1}^N \sum_{j=1}^{2^K} w_n[j] q_n[j]^2 \leq \frac{\ln(2^K)}{\mu} + \frac{\mu}{2}\sum_{n=1}^N l_{\max}^2 \nonumber\\
& = \frac{\ln(2^K)}{\mu} + \frac{\mu N l_{\max}^2}{2} \leq l_{\max}\sqrt{\ln(S) N}.
\end{align} Note that $S = 2^{K+1}-1$. The above inequality holds for any $w^*\in \Delta(2^K)$, including the $w^e$ that corresponds to the expert (i.e., $w^e[1] = 1, w^e[i]=0,i\neq 1$ as we assumed without loss of generality the left most trajectory is the optimal trajectory).

Now let us define $\ell_n(w)$ as follows:
\begin{align}
\ell_n(w) = \frac{1}{K+1}\sum_{t=1}^{K+1}\sum_{s\sim \mathcal{S}} d_t^{\pi_n}(s)\sum_{j=1}^{2^K} w_j Q^*(s,\pi^j(s)).
\end{align} Note $\ell_n(w)$ can be understood as first rolling in $\pi_n$ \emph{infinitely many times} and then querying for the exact cost-to-go $Q^*$ on all the visited states. Clearly $\tilde{\ell}_n(w)$ is an unbiased estimate of $\ell_n(w)$: $\mathbb{E}[\tilde{\ell}_n(w)] -\ell_{n}(w) = 0$, where the expectation is over the randomness of the roll-in and sampling procedure of $\tilde{Q}^e$ at iteration $n$, conditioned on all events among the previous $n-1$ iterations. Also note that $|\tilde{\ell}_n(w) - \ell_n(w)| \leq 2l_{\max}$, since $\ell_n(w) \leq l_{\max}$. Hence $\{\tilde{\ell}_n(w_n) - \ell_n(w_n)\}$ is a bounded martingale difference sequence. Hence by Azuma-Heoffding inequality, we get with probability at least $1-\delta/2$:
\begin{align}
\sum_{n=1}^{N} {\ell}_n(w_n) - \tilde{\ell}_n(w_n) \leq 2l_{\max}\sqrt{\log(2/\delta)N},
\end{align} and with probability at least $1-\delta/2$:
\begin{align}
\sum_{n=1}^{N} \tilde{\ell}_n(w^e) - {\ell}_{n}(w^e) \leq 2l_{\max}\sqrt{\log(2/\delta)N}.
\end{align} Combine the above inequality using union bound, we get with probability at least $1-\delta$:
\begin{align}
\sum_{n=1}^N (\ell_n(w_n) - \ell_n(w^e)) \leq\sum_{n=1}^N (\tilde{\ell}_n(w_n) - \tilde{\ell}_n(w^e)) + 4l_{\max}\sqrt{\log(2/\delta)N}. 
\end{align}

Now let us apply the Performance Difference Lemma (Lemma~\ref{lemma:performance_difference}),  we get with probability at least $1-\delta$:
\begin{align}
\sum_{n=1}^N \mu(\pi_n) - \sum_{n=1}^N \mu(\pi^*) = \sum_{n=1}^N (K+1) \big(\ell_n(w_n) - \ell_n(w^e)\big)  \leq (K+1)(l_{\max}\sqrt{\ln(S)N} +4l_{\max}\sqrt{\log(2/\delta)N}),
\end{align} rearrange terms we get:
\begin{align}
\sum_{n=1}^N \mu(\pi_n) - \sum_{n=1}^N \mu(\pi^*) \leq \log(S)l_{\max}(\sqrt{\ln(S)N} + \sqrt{\log(2/\delta)N}) \leq O(\ln(S)\sqrt{\ln(S)N}),
\end{align} with probability at least $1-\delta$.
\end{proof}

\section{Proof of Theorem~\ref{them:upper_bound}}
\label{sec:proof_upper_bound}
The proof of theorem~\ref{them:upper_bound} is similar to the one for theorem~\ref{them:special_lower_noisy}. Hence we simply consider the infinitely many roll-ins and exact query of $Q^*$ case. The finite number roll-in and noisy query of $Q^*$ case can be handled by using the martingale difference sequence argument as shown in the proof of theorem~\ref{them:special_lower_noisy}.

\begin{proof}
Recall that in general setting, the policy $\pi$ consists of probability vectors $\pi^{s,t}\in\Delta(A)$, for all $s\in\mathcal{S}$ and $t\in[H]$: $\pi = \{\pi^{s,t}\}_{\forall s\in \mathcal{S},t\in[H]}$. Also recall that the loss functions EG is optimizing are $\{\ell_n(\pi)\}$ where:
\begin{align}
\ell_n(\pi) = \frac{1}{H}\sum_{t=1}^H\sum_{s\in\mathcal{S}} d_t^{\pi_n}(s) (\pi^{s,t}\cdot Q_t^*(s)) = \sum_{t=1}^H\sum_{s\in\mathcal{S}}\pi^{s,t}\cdot q_n^{s,t}
\end{align} where as we defined before $Q_t^*(s)$ stands for the cost-to-go vector $Q_t^*(s)[j] = Q_t^*(s,a_j)$, for the $j$'th action in $\mathcal{A}$, and $q_n^{s,t} = \frac{d_t^{\pi_n}(s)}{H}Q_t^*(s)$.

Now if we run Exponential Gradient descent on $\ell_n$ to optimize $\pi^{s,t}$ for each pair of state and time step independently, we can get the following regret upper bound by using Lemma~\ref{lemma:EG}:
\begin{align}
\sum_{n=1}^N \ell_n(\pi) - \min_{\pi}\sum_{n=1}^N\ell_n(\pi_n) \leq \sum_{t=1}^H\sum_{s\in\mathcal{S}}\big( \frac{\ln(A)}{\mu} + \frac{\mu}{2}\sum_{n=1}^N\sum_{j=1}^A \pi^{s,t}[j] q_n^{s,t}[j]^2\big).
\end{align} Note that we can upper bound $(q_n^{s,t}[j])^2$ as:
\begin{align}
(q_n^{s,t}[j])^2 \leq \frac{d_t^{\pi_n}(s)^2}{H^2}(Q^*_{\max})^2 \leq \frac{d_t^{\pi_n}(s)}{H^2} (Q^2_{\max})^2 
\end{align}
Substitute it back, we get:
\begin{align}
&\sum_{n=1}^N (\ell_n(\pi_n) - \ell_n(\pi^*)) \leq \sum_{t=1}^H\sum_{s\in\mathcal{S}} \big(\frac{\ln(A)}{\mu} + \frac{\mu}{2}\sum_{n=1}^N\sum_{j=1}^A \pi^{s,t}[j] d_t^{\pi_n}(s)\frac{(Q^*_{\max})^2}{H^2} \big) \nonumber\\
& = \sum_{t=1}^H \big(\frac{S\ln(A)}{\mu} + \frac{\mu(Q^*_{\max})^2}{2H^2}\sum_{n=1}^N\sum_{s\in\mathcal{S}}d_t^{\pi_n}(s)\sum_{j=1}^A\pi^{s,t}[j]\big) = \sum_{t=1}^H \big( \frac{S\ln(A)}{\mu} + \frac{\mu(Q^*_{\max})^2}{2H^2}N\big) \nonumber\\
& \leq \frac{Q^*_{\max}}{H}\sqrt{2S\ln(A)N},
\end{align} if we set $\mu = \sqrt{(Q^*_{\max})^2NS\ln(A)/(2H^2)}$.

Now let us apply the performance difference lemma (Lemma~\ref{lemma:performance_difference}), we get:
\begin{align}
R_N = \sum_{n=1}^N\mu(\pi_n) - \sum_{n=1}^N\mu(\pi^*) = H\sum_{n=1}^N (\ell_n(w_n) - \ell_n(w^e)) \leq HQ_{\max}^e\sqrt{S\ln(A)N}.
\end{align}

\end{proof}

\section{Proof of Theorem~\ref{them:lower_bound}}
\label{sec:proof_lower_bound}
Let us use $\tilde{Q}^e(s)$ to represent the noisy but unbiased estimate of $Q^*(s)$.
\begin{proof}
For notation simplicity, we denote $\mathcal{S} = \{s_1, s_2, ..., s_S\}$. We consider a finite MDP with time horizon $H = 1$. The initial distribution $\rho_0 = \{1/S, ..., 1/S\}$ puts $1/S$ weight on each state. 
We consider the algorithm setting where at every episode $n$, a state $s^n\in \mathcal{S}$ is sampled from $\rho_0$ and the algorithm uses its current policy $\pi_{n}^{s_n}\in\Delta({A})$ to pick an action $a\in\mathcal{A}$ for $s^n$ and then receives a noisy but unbiased estimate $\tilde{Q}^e(s^n)$ of $Q^*(s^n)\in\mathbb{R}^{|\mathcal{A}|}$. The algorithm then updates its policy from $\pi_{n}^{s^n}$ to $\pi_{n+1}^{s^n}$ for $s^n$ while keep the other polices for other $s$ unchanged (since the algorithm did not receive any feedback regarding $Q^*(s)$ for $s\neq s^{n}$ and the sample distribution $\rho_0$ is fixed and uniform). For expected regret $\mathbb{E}[R_N]$ we have the following fact:
\begin{align}
\label{eq:relation}
&\mathop{\mathbb{E}}_{s^n\sim\rho_0,\forall n}\Big[\mathop{\mathbb{E}}_{\tilde{Q}^e(s_n)\sim P_{s_n},\forall n}\big[\sum_{n=1}^N( \pi_n^{s^n}\cdot \tilde{Q}^e(s^n) - \pi^*_{s^n}\cdot \tilde{Q}^e(s^n))\big]\Big] \nonumber\\
&= \mathop{\mathbb{E}}_{s^n\sim \rho_0,\forall n}\Big[\sum_{n=1}^N\mathop{\mathbb{E}}_{\tilde{Q}^e_i(s_i)\sim P_{s_i},i\leq n-1}\big[ (\pi_n^{s^n}\cdot Q^*(s^n) - \pi_{s^n}^e\cdot Q^*(s^n))\big]\Big] \nonumber\\
& = \sum_{n=1}^N\mathop{\mathbb{E}}_{s^i\sim \rho_0,i\leq n-1}\Big[\mathop{\mathbb{E}}_{\tilde{Q}^e_i(s_i)\sim P_{s_i},i\leq n-1}\big[ \mathop{\mathbb{E}}_{s\sim\rho_0}(\pi_n^{s}\cdot Q^*(s) - \pi_{s}^*\cdot Q^*(s))\big]\Big] \nonumber\\
&= \mathbb{E}\big[\sum_{n=1}^N\mathop{\mathbb{E}}_{s\sim\rho_0}\pi_n^s\cdot Q^*(s) - \mathop{\mathbb{E}}_{s\sim \rho_0}\pi_s^*\cdot Q^*(s)\big] \nonumber\\
& = \mathbb{E}\sum_{n=1}^N [\mu(\pi_n) - \mu(\pi^*)],
\end{align}where the expectation in the final equation is taken with respect to random variables $\pi_i,i\in[N]$ since each $\pi_i$ is depend on $\tilde{Q}^e_j$, for $j< i$ and $s^j$, for $j<i$.  

We first consider $\mathop{\mathbb{E}}_{\tilde{Q}^e(s^n)\sim P_{s^n},\forall n}\big[\sum_{n=1}^N(  \pi_n^{s^n}\cdot \tilde{Q}^e(s^n) - \pi^*_{s^n}\cdot \tilde{Q}^e(s^n))\big]$ conditioned on a given sequence of $s^1,...,s^N$.
Let us define that among $N$ episodes, the set of the index of the episodes that state $s_i$ is sampled as $\mathcal{N}_i$ and its cardinality as $N_i$, and we then have $\sum_{i=1}^S N_i = N$ and $\mathcal{N}_i \cap\mathcal{N}_j = \emptyset$,for $i\neq j$. 
\begin{align}
&\mathop{\mathbb{E}}_{\tilde{Q}^e(s^n)\sim P_{s^n},\forall n}\big[\sum_{n=1}^N(  \pi_n^{s^n}\cdot \tilde{Q}^e(s^n) - \pi^*_{s^n}\cdot \tilde{Q}^e(s^n))\big] \nonumber\\
& = \sum_{i=1}^S \sum_{j\in \mathcal{N}_i} \mathop{\mathbb{E}}_{\tilde{Q}_j^e(s_i)\sim P_{s_i}}(\pi_{j}^{s_i}\cdot\tilde{Q}_j^e(s_i) - \pi_{s_i}^e\tilde{Q}_j^e(s_i))
\label{eq:composed}
\end{align}

Note that for each state $s_i$, at the rounds from $\mathcal{N}_i$, we can think of the algorithm running any possible online linear regression algorithm to compute the sequence of policies $\pi_j^{s_i},\forall j\in \mathcal{N}_i$ for state $s_i$. Note that from classic online linear regression analysis, we can show that for state $s_i$ there exists a distribution $P_{s_i}$ such that for any online algorithm:
\begin{align}
\mathop{\mathbb{E}}_{\tilde{Q}^e_j(s_i)\sim P_{s_i},\forall j\in\mathcal{N}_i}\big[ \sum_{j\in\mathcal{N}_i} (\pi_{j}^{s_i} \cdot \tilde{Q}_j^e(s_i) - \pi_{s_i}^e\cdot \tilde{Q}_j^e(s_i))  \big] \geq c\sqrt{\ln(A) N_i},
\end{align} for some non-zero positive constant $c$.
Substitute the above inequality into Eq.~\ref{eq:composed}, we have:
\begin{align}
&\mathop{\mathbb{E}}_{\tilde{Q}^e(s_n)\sim P_{s_n},\forall n}\big[\sum_{n=1}^N(   \pi_n^{s^n}\cdot \tilde{Q}^e(s^n) -\pi^*_{s^n}\cdot \tilde{Q}^e(s^n))\big]\geq \sum_{i=1}^S c\sqrt{\ln(A)N_i} = c\sqrt{\ln(A)}\sum_{i=1}^S\sqrt{N_i}.
\end{align}
Now let us put the expectation $\mathbb{E}_{s^i\sim\rho_0,\forall i}$ back, we have:
\begin{align}
\label{eq:fact_1}
&\mathop{\mathbb{E}}_{s^n\sim\rho_0,\forall n}\Big[\mathop{\mathbb{E}}_{\tilde{Q}^e(s_n)\sim P_{s_n}}\big[\sum_{n=1}^N( \pi_n^{s^n}\cdot \tilde{Q}^e(s^n) - \pi^*_{s^n}\cdot \tilde{Q}^e(s^n))|s^1,...,s^n\big]\Big]\geq  c\sqrt{\ln(A)}\sum_{i=1}^N\mathbb{E}[\sqrt{N_i}].
\end{align}

Note that each $N_i$ is sampled from a Binomial distribution $\mathcal{B}(N, 1/S)$. To lower bound $\mathbb{E}_{n\sim \mathcal{B}(N,1/S)} \sqrt{n}$, we use Hoeffding's Inequality here.  Note that $N_i = \sum_{n=1}^N a_n$, where $a_n = 1$ if $s_i$ is picked at iteration $n$ and zero otherwise. Hence $a_i$ is from a Bernoulli distribution with parameter $1/S$. Using Hoeffding bound, for $N_i/N$, we get:
\begin{align}
P(|N_i/N - 1/S| <= \epsilon) \geq 1 - \exp(-2N\epsilon^2).
\end{align} Let $\epsilon = 1/(2S)$, and substitute it back to the above inequality, we get:
\begin{align}
P(0.5(N/S)\leq N_i \leq 1.5(N/S)) = P(\sqrt{0.5(N/S)}\leq \sqrt{N_i} \leq \sqrt{1.5(N/S)})\geq 1 - \exp(-2N/S^2).
\end{align}
Hence, we can lower bound $\mathbb{E}[\sqrt{N_i}]$ as follows:
\begin{align}
\mathbb{E}[\sqrt{N_i}] \geq \sqrt{0.5N/S}(1 - \exp(-2N/S^2)).
\end{align} Take $N$ to infinity, we get:
\begin{align}
\lim_{N\to\infty}\mathbb{E}[\sqrt{N_i}] \geq \sqrt{0.5N/S}.
\end{align}
Substitute this result back to Eq.~\ref{eq:fact_1} and use the fact from Eq.~\ref{eq:relation}, we get:
\begin{align}
\lim_{N\to\infty}\mathbb{E}[R_N] = \lim_{N\to\infty}&\mathop{\mathbb{E}}_{s^n\sim\rho_0,\forall n}\Big[\mathop{\mathbb{E}}_{\tilde{Q}^e(s_n)\sim P_{s_n},\forall n}\big[\sum_{n=1}^N( \pi_n^{s^n} \cdot \tilde{Q}^e(s^n) - \pi^*_{s^n}\cdot \tilde{Q}^e(s^n))\big]\Big]\geq c\sqrt{\ln(A)}\sum_{i=1}^S\mathbb{E}[\sqrt{N_i}] \nonumber\\
&\geq c\sqrt{\ln(A)}S\sqrt{0.5 N/S} = \Omega(\sqrt{S\ln(A)N}). \nonumber
\end{align}
Hence we prove the theorem.
\end{proof}


\section{Details of Dependency Parsing for Handwritten Algebra}
\label{sec:parsing_example}
\begin{figure}[h]
	\centering
	\vspace{-2mm}
	\begin{subfigure}[l]{0.45\textwidth}
        \includegraphics[width=1.1\textwidth,keepaspectratio]{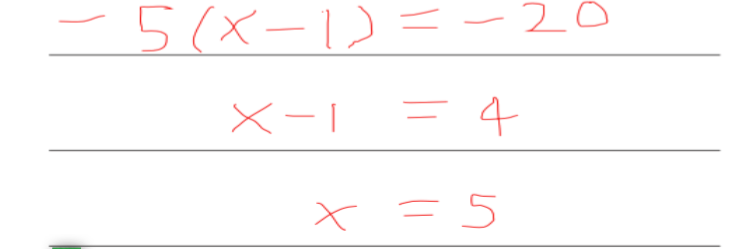}
        \caption{Handwritten algebra equations}
        \label{fig:cartpole}
    \end{subfigure}
	\begin{subfigure}[l]{0.5\textwidth}
        \includegraphics[width=1.1\textwidth,keepaspectratio]{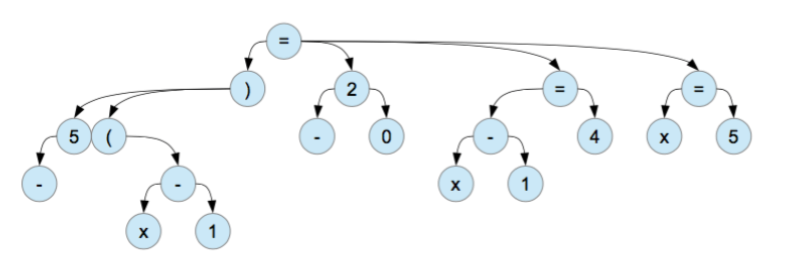}
        \caption{Dependency tree}
        \label{fig:acrobot}
    \end{subfigure}
    \caption{An example of a set of handwritten algebra equations (a) and its corresponding dependency tree (b).}
    \label{fig:algebra_example}
\end{figure}

In Fig.~\ref{fig:algebra_example}, we show an example of set of handwritten algebra equations and its dependency tree from a arc-hybird sequence $slssslssrrllslsslssrrslssrlssrrslssrr$.  The preprocess step cropped individual symbols one by one from left to right and from the top equation to the bottom one, centered them, scaled symbols to 40 by 40 images,  and  finally formed  them  as  a  sequence of images. 

Since in the most common dependency parsing setting, there is no immediate reward at every parsing step,
the reward-to-go $Q^*(s,a)$ is computed by using UAS as follows: start from $s$ and apply action $a$, then use expert $\pi^*$ to roll out til the end of the parsing process; $Q^*(s,a)$ is the UAS score of the final configuration. Hence AggreVaTeD can be considered as directly maximizing the UAS score, while previous approaches such as DAgger or SMILe \cite{Ross2011_AISTATS} tries to mimic expert's actions and hence are not directly optimizing the final objective.

\end{document}